\documentclass{article}

\usepackage{microtype}
\usepackage{graphicx}
\usepackage{booktabs} %

\usepackage[pagebackref]{hyperref}

\usepackage[accepted]{icml2024}

\usepackage{amsmath}
\usepackage{amssymb}
\usepackage{mathtools}
\usepackage{amsthm}

\usepackage[ruled,linesnumbered,vlined,algo2e]{algorithm2e}

\usepackage[capitalize,noabbrev]{cleveref}

\usepackage[textsize=tiny]{todonotes}

\usepackage{thmtools} 
\usepackage{thm-restate}

\usepackage{listings}
\usepackage{caption}
\usepackage{subcaption}

\usepackage{macro}
\usepackage[normalem]{ulem}
\usepackage{enumitem}

\usepackage[frozencache]{minted}

\newcommand{\pderivative}[2][]{\frac{\partial#1}{\partial#2}}
\newcommand{\derivative}[2][]{\frac{\diff#1}{\diff#2}}

\icmltitlerunning{Understanding Stochastic Natural Gradient Variational Inference}

\newcommand{\jrg}[1]{}
\newcommand{\kw}[1]{}

\begin{document}

\twocolumn[
\icmltitle{Understanding Stochastic Natural Gradient Variational Inference}

\icmlsetsymbol{equal}{*}

\begin{icmlauthorlist}
\icmlauthor{Kaiwen Wu}{upenn}
\icmlauthor{Jacob R. Gardner}{upenn}
\end{icmlauthorlist}

\icmlaffiliation{upenn}{Department of Computer and Information Science, University of Pennsylvania, Philadelphia, United States}

\icmlcorrespondingauthor{Kaiwen Wu}{kaiwenwu@seas.upenn.edu}
\icmlcorrespondingauthor{Jacob R. Gardner}{jacobrg@seas.upenn.edu}

\icmlkeywords{Machine Learning, ICML}

\vskip 0.3in
]

\printAffiliationsAndNotice{}  %

\begin{abstract}
Stochastic natural gradient variational inference (NGVI) is a popular posterior inference method with applications in various probabilistic models.
Despite its wide usage, little is known about the non-asymptotic convergence rate in the \emph{stochastic} setting.
We aim to lessen this gap and provide a better understanding.
For conjugate likelihoods, we prove the first $\mathcal{O}(\frac{1}{T})$ non-asymptotic convergence rate of stochastic NGVI.
The complexity is no worse than stochastic gradient descent (\aka black-box variational inference) and the rate likely has better constant dependency that leads to faster convergence in practice.
For non-conjugate likelihoods, we show that stochastic NGVI with the canonical parameterization implicitly optimizes a non-convex objective.
Thus, a global convergence rate of $\mathcal{O}(\frac{1}{T})$ is unlikely without some significant new understanding of optimizing the ELBO using natural gradients.
\end{abstract}

\section{Introduction}

Given a prior $p(\zv)$ and a likelihood $p(\yv \mid \zv)$, variational inference (VI) approximates the posterior $p(\zv \mid \yv)$ by optimizing the evidence lower bound (ELBO) in a family of variational distributions \citep{blei2017variational}.
Natural gradient variational inference (NGVI), in particular, optimizes the ELBO by natural gradient descent (NGD) \citep{amari1998natural}.\jrg{What is the posterior? What is the ELBO?}\kw{added a citation to Blei et al. to explain the ELBO}\jrg{"the gradient descent" $\to$ "gradient descent." "the natural gradient descent" $\to$ "natural gradient descent,"}
Different from (standard) gradient descent that follows the steepest descent direction induced by the Euclidean distance, NGD follows the steepest descent direction induced by the KL divergence \cite{honkela2004unsupervised,hensman2012fast,hoffman2013stochastic}.\jrg{We've fully described NGD in text here, but there are is only one citation to a 1998 NGD paper.}\kw{more citations added}\kw{more citations added}\kw{more citations added}
The folk wisdom is that the KL divergence is a better ``metric'' to compare distributions and thus NGD is believed to be superior than gradient descent, \aka black-box variational inference \citep{ranganath2014black}.
Indeed, NGVI as well as its variants empirically outperforms gradient descent in many cases, and thus enjoys applications in a wide range of probabilistic models.
Here, we name a few exmaples: latent Dirichlet allocation topic models \citep{hoffman2013stochastic}, Bayesian neural networks \citep{khan2018fast,osawa2019practical}, and
large-scale Gaussian processes \citep{hensman2013gaussian,hensman2015scalable,salimbeni2018natural}.

Despite its wide usage, a non-asymptotic convergence rate of NGVI in the stochastic setting is absent, even for simple conjugate likelihoods.
A few convergence arguments exist in the literature, but none of them applies to any practical uses of NGVI.
For example, \citet{hoffman2013stochastic} have a convergence argument
\footnote{\citet{hoffman2013stochastic} did not give a convergence proof besides a reference to \citet{bottou1998online}.}
by assuming the Fisher information matrix has eigenvalues bounded from below (by a positive constant) throughout the natural gradient updates.
\citet{khan2016faster} analyze a variant of NGVI based on Bregman proximal gradient descent by assuming the (KL) divergence is $\alpha$-strongly convex, a condition that generally does not hold (at least for the KL divergence).
Besides, \citet{khan2016faster} did not obtain a complexity bound in the stochastic setting---they only showed convergence to a region around stationary points.
Note that these assumptions do not hold in the entire domain, provably.
Even if they hold in a subset of the domain, the constants in these assumptions are difficult to estimate, and might even be arbitrarily bad as the posterior distribution $p(\zv \mid \yv)$ contracts.\footnote{For instance, the Fisher information matrix gets increasingly close to singular as the covariance of the posterior $p(\zv \mid \yv)$ shrinks.}

\jrg{This discussion about us vs them needs to appear at least somewhat in the intro. Nowhere in the intro do we cite that *any* theoretical work exists on stochastic NGVI}
\kw{Moved part of the argument to the introduction.}

This work aims to lessen this gap and obtain a ``clean'' analysis, with minimal assumptions, that is applicable to some practical uses of stochastic NGVI.
For the sake of generality, existing analyses have to use assumptions that does not hold in practice.
Therefore, we pursue the opposite direction of generality---the basic setting of conjugate likelihoods, for which we establish the first $\Oc(\frac{1}{T})$ non-asymptotic convergence rate of stochastic natural gradient variational inference.
This rate has the same complexity as the convergence rate of stochastic projected (and proximal) gradient descent recently studied by \citet{domke2020provable,domke2023provable,kim2023convergence}.
This, along with our experiments, implies that NGVI ultimately may share the same complexity with other first-order methods.
The empirical observation that NGVI is faster than stochastic gradient descent is likely due to a better constant dependency in the big $\Oc$ notation.
Indeed, as we will see later, our convergence rate of stochastic NGVI is independent of the objective's condition number and the distance from the initialization to the optimum.
Nevertheless, the constant improvement may play a huge difference in practice.
\jrg{This is more imprecise than your paper deserves. You have some specific ideas of \textit{how} the constant is better. Discuss this.}
\kw{added a sentence describe why the constant is better}

Although our convergence rate for stochastic NGVI assumes conjugate likelihoods, it is already applicable to some practical uses, including large-scale Bayesian linear regression and variational parameter learning in stochastic variational Gaussian processes \citep{hensman2013gaussian,hensman2015scalable,salimbeni2018natural}.
Indeed, we will show that all assumptions are strictly satisfied in practice and the constant in the convergence rate can be bounded explicitly using statistics from the training data.\jrg{Please read the Salimbeni et al., paper carefully. They argue that NGD is very useful specifically in the non-conjugate case.
This is in contrast to your own theoretical results. Discuss this.}\kw{added a discussion in the end of the next paragraph}

For non-conjugate likelihoods, we show that the ``canonical'' implementation of stochastic NGVI implicitly optimizes a non-convex objective even when the likelihoods are simple log-concave distributions.
Hence, the convergence behavior of stochastic NGVI with non-conjugate likelihoods is more nuanced, which might partially explain why the theoretical understanding of stochastic NGVI is lacking throughout the years.
This lack of convexity implies that proving a global convergence rate of $\Oc(\frac1T)$ for non-conjugate likelihoods may require new properties of the ELBO, \eg, the Polyak-{\L}ojasiewicz inequality \citep{polyak1963gradient,lojasiewicz1963topological}, in order to explain the empirical success of stochastic NGVI for non-conjugate likelihoods \citep[\eg,][]{hoffman2013stochastic,salimbeni2018natural}.
\kw{cited}

\section{Background}

\textbf{Notation.}
We use $\norm{\cdot}$ to denote the vector Euclidean norm.
For matrices, the same symbol $\norm{\cdot}$ is overloaded to denote the spectral norm.
$\norm{\cdot}_\F$ denotes the Frobenius norm.
$\inner{\cdot, \cdot}$ denotes an inner product, whose domain is inferred from its arguments.
Let $\Ds_\KL(\cdot, \cdot)$ denote the Kullback–Leibler divergence between distributions.
$\Sb_{++}^d$ (and $\Sb_{+}^d$) represents the collection of all $d \times d$ symmetric positive (semi-)definite matrices.
Let $\succ$ (and $\succeq$) be the partial order induced by $\Sb_{++}^d$ (and $\Sb_{+}^d$), \ie, $\Av \succ \Bv$ if and only if $\Av - \Bv \in \Sb_{++}^d$.

\subsection{Variational Inference with Exponential Families}

Suppose we have a prior $p(\zv)$ on latent variables $\zv$ and a likelihood $p(\yv \mid \zv)$ on observations $\yv$.
Variational inference (VI) aims to find the best approximation of the posterior $p(\zv \mid \yv)$ inside a variational family $\Qc$ by minimizing the Kullback–Leibler (KL) divergence
\begin{align*}
    \mini_{q \in \Qc} \Ds_{\KL}\bb{q(\zv), p(\zv \mid \yv)},
\end{align*}
where $q$ is the variational distribution.
This is the equivalent to minimizing the objective
\begin{align}
\label{eq:elbo}
    \ell(q) = -\Eb_{q(\zv)}[\log p(\yv \mid \zv)] + \Ds_{\KL}(q(\zv), p(\zv)),
\end{align}
which is called the negative evidence lower bound (ELBO).
Throughout the paper, we assume the variational family $\Qc$ is an exponential family (which will be defined below), and the prior $p(\zv)$ is in $\Qc$.
Though, the posterior $p(\zv \mid \yv)$ is not necessarily in $\Qc$, unless the likelihood is conjugate: we call the likelihood $p(\yv \mid \zv)$ conjugate (with the prior) if and only if $p(\zv \mid \yv) \in \Qc$. 
Conjugacy implies the variational approximation is exact, so long as \eqref{eq:elbo} is minimized globally.

\textbf{Exponential Family.}
A (regular and minimal) exponential family is a collection of distributions indexed by a canonical parameter $\etav$ in the form
\begin{align}
\label{eq:exponential-family}
    q(\zv; \etav) = h(\zv) \exp\big(\inner{\phi(\zv), \etav} - A(\etav)\big),
\end{align}
where $h$ is the base measure, $\phi$ is the sufficient statistic, $\etav$ is the natural parameter, and $A$ is the log-partition function.

The set of all possible $\etav$ that make $q(\zv; \etav)$ integrable forms an open convex set $\Dc$, called the natural parameter space.
The log-partition function $A: \Dc \to \Rb$ is differentiable and strictly convex on $\Dc$.
The associated expectation parameter $\omegav$ of $q(\zv; \etav)$ is defined as the expected sufficient statistic:
\begin{align}
\label{eq:expectation-parameter}
    \omegav = \Eb_{q(\zv; \etav)} \sbb{\phi(\zv)}.
\end{align}
The set of all possible expectation parameters $\omegav$ again forms a convex set $\Omega$, called the expectation parameter space.

The natural and expectation parameter spaces, $\Dc$ and $\Omega$, are linked by the gradients of the log-partition function $A$ and its convex conjugate $A^*$, where the differentiable and strictly convex function $A^*: \Omega \to \Rb$ is defined as
\begin{align*}
    A^*(\omegav) = \max_{\etav \in \Dc} \inner{\etav, \omegav} - A(\etav).
\end{align*}
Indeed, the gradient maps $\nabla A: \Dc \to \Omega$ and $\nabla A^*: \Omega \to \Dc$ are inverses of each other.
Namely, if $\etav \in \Dc$ and $\omegav \in \Omega$ satisfy \eqref{eq:expectation-parameter} representing the same distribution, then
\begin{align}
\label{eq:mirror-map}
    \nabla A(\etav) = \omegav, \quad \nabla A^*(\omegav) = \etav.
\end{align}

\textbf{Example.}
A $d$-dimensional Gaussian distribution $\Nc(\muv, \Sigmav)$ has its natural parameter $\etav = (\lambdav, \Lambdav)$ defined on
\begin{align}
\label{eq:natural-parameter-domain}
    \Dc = \cbb{\etav = (\lambdav, \Lambdav) \in \Rb^d \times \Rb^{d \times d}: -\Lambdav \in \Sb_{++}^d}
\end{align}
with the parameter conversion identity $\lambdav = \Sigmav^{-1} \muv$ and $\Lambdav = -\frac12 \Sigmav^{-1}$.
The expectation parameter $\omegav$ is defined on
\begin{align}
\label{eq:expectation-parameter-domain}
    \Omega = \cbb{\omegav = (\xiv, \Xiv) \in \Rb^d \times \Rb^{d \times d}: \Xiv - \xiv \xiv^\top \in \Sb_{++}^d}
\end{align}
with the identity $\xiv = \muv$ and $\Xiv = \Sigmav + \muv \muv^\top$.
See \S\ref{sec:exponential-family} for more details, where we give explicit expressions for $A(\etav)$, $A^*(\omegav)$, $\nabla A(\etav)$ and $\nabla A^*(\omegav)$.
For more background on exponential families, we direct readers to the monograph by \citet{wainwright2008graphical}.

\clearpage

\begin{remark}[Overload $\ell$]
Let $\etav$ and $\omegav$ be the natural and expectation parameters of the same distribution $q$.
We have
\begin{align*}
    \ell(q) = \ell^{(\mathrm{n})}(\etav) = \ell^{(\mathrm{e})}(\omegav),
\end{align*}
where $\ell^{(\mathrm{n})}$ and $\ell^{(\mathrm{e})}$ are the negative ELBO as functions of the natural and expectation parameters respectively.
Technically, $\ell$, $\ell^{(\mathrm{n})}$ and $\ell^{(\mathrm{e})}$ are different functions with different domains and arguments.
For notation simplicity, however, we will drop the superscript when the context allows---the superscript will be inferred from the argument.
\end{remark}

\subsection{Natural Gradient Descent for Variational Inference}
Natural gradient variational inference (NGVI) optimizes the ELBO by natural gradient descent (NGD).
It iteratively updates the natural parameter $\etav$ of the variational distribution by taking a steepest descent step induced by the KL divergence.
The yielded update rule is a preconditioned gradient descent with the Fisher information matrix (FIM).
\begin{definition}[FIM]
Given a (not necessarily exponential family) distribution $q(\zv; \etav)$ parameterized by $\etav$, the Fisher information matrix is defined as
\begin{align*}
    \Fv\bb{\etav} = -\Eb_{q(\zv)} [\nabla_{\etav}^2 \log q(\zv; \etav)],
\end{align*}
where $\nabla^2$ is taken \wrt $\etav$.
In particular, for the exponential family \eqref{eq:exponential-family}, it takes a simple form $\Fv(\etav) = \nabla^2 A(\etav)$.
\end{definition}
\begin{definition}[NGD]
Natural gradient descent iterates
\begin{align}
\label{eq:natural-gradient-update}
    \etav_{t + 1} = \etav_t - \gamma_t \Fv\bb{\etav_t}^{-1} \nabla \ell(\etav_t),
\end{align}
where $\gamma_t$ is the step size and $\Fv\bb{\etav}$ is the Fisher information matrix.
The FIM-preconditioned gradient $\Fv(\etav_t)\inv \nabla \ell(\etav_t)$ is often called the ``natural'' gradient.
\end{definition}
We call the update \eqref{eq:natural-gradient-update} the ``canonical'' NGD update, as NGD can be implemented in other parameters beyond the natural parameter $\etav$.
However, typically NGD converges the fastest in the natural parameterization, which will be the main focus of this paper.
We will revisit other parameterizations in \S\ref{sec:landscape}.

Explicitly inverting the FIM is inefficient, \eg, its takes $\Oc(d^6)$ time for a Gaussian due to its $d(d + 1) \times d(d + 1)$ size.
Fortunately, the NGD update can be implemented without explicit FIM inversion for the exponential family \citep{raskutti2015information}.
Let $\etav$ and $\omegav$ be the natural and expectation parameters of the same distribution, hence having the same ELBO value $\ell^{(\mathrm{n})}(\etav) = \ell^{(\mathrm{e})}(\omegav)$. 
Plugging in the identity $\omegav = \nabla A(\etav)$ as in \eqref{eq:mirror-map}, we obtain
\begin{align*}
    \ell^{(\mathrm{n})}(\etav) = \ell^{(\mathrm{e})}(\nabla A(\etav)).
\end{align*}
Differentiating \wrt $\etav$ on both sides gives
\begin{align*}
    \nabla \ell^{(\mathrm{n})}(\etav) & = \nabla^2 A(\etav) \cdot \nabla \ell^{(\mathrm{e})} (\nabla A(\etav)) \\
    & = \Fv(\etav) \nabla \ell^{(\mathrm{e})} (\omegav),
\end{align*}
which implies $\Fv(\etav)^{-1} \nabla \ell^{(\mathrm{n})}(\etav) = \nabla \ell^{(\mathrm{e})} (\omegav)$: the natural gradient (of the natural parameter) is simply the gradient of the (negative) ELBO \wrt the expectation parameter.
Thus, the NGD update rule \eqref{eq:natural-gradient-update} reduces to
\begin{align}
\label{eq:natural-gradient-update-inverse-free}
    \etav_{t + 1} = \etav_t - \gamma_t \nabla \ell(\omegav_t),
\end{align}
with no explicit FIM inversion.

\subsection{Natural Gradient Descent as Mirror Descent}
This section reviews the connection between NGD and  mirror descent (MD).
This connection was discovered by \citet{raskutti2015information} and later applied to variational inference by \citet{khan2017conjugate,khan2018fast}.

\begin{definition}[Bregman Divergence]
Given a differentiable and strictly convex function $\Phi$, the associated Bregman divergence is defined as
\begin{align*}
    \Ds_\Phi(\av, \bv) = \Phi(\av) - \Phi(\bv) - \inner{\nabla \Phi(\bv), \av - \bv},
\end{align*}
and we call $\Phi$ the distance generating function.
\end{definition}
Recall that $A^*$, as the convex conjugate of the log-partition function $A$, is differentiable and strictly convex on $\Omega$.
Thus, it is a valid distance generating function and induces a Bregman divergence $\Ds_{A^*}$ on the expectation parameters.
The divergence is then used to define mirror descent \citep{nemirovskij1983problem}, which iteratively solves regularized first-order approximations.
\begin{definition}[MD]
\label{def:mirror-descent}
The mirror descent update is defined as
\begin{align}
\label{eq:mirror-descent-update-proximal-form}
    \omegav_{t + 1} = \argmin_{\omegav \in \Omega} \inner{\nabla \ell(\omegav_t), \omegav} + \frac{1}{\gamma_t} \Ds_{A^*}(\omegav, \omegav_t),
\end{align}
where $\gamma_t > 0$ is the step size.
\end{definition}
Mirror descent (MD) is a generalization of gradient descent.
If the Bregman divergence $\Ds_{A^*}(\omegav, \omegav_t)$ in \eqref{eq:mirror-descent-update-proximal-form} is replaced with the squared Euclidean norm $\frac12 \norm{\omegav - \omegav_t}^2$, we recover the familiar update rule $\omegav_{t+1} = \omegav_t - \gamma_t \nabla \ell(\omegav_t)$.

To implement MD efficiently, the minimization in \eqref{eq:mirror-descent-update-proximal-form} needs to be solved in closed-form.
Taking the derivative \wrt $\omegav$ on both sides and setting it equal to zero, we obtain
\begin{align}
\label{eq:mirror-descent-update-dual-space-form}
    \nabla A^*(\omegav_{t + 1}) & = \nabla A^*(\omegav_t) - \gamma_t \nabla \ell(\omegav_t).
\end{align}
Recall $\nabla A^*: \Omega \to \Dc$ maps the expectation parameter $\omegav$ of an exponential family distribution $q$ to its natural parameter $\etav$, \ie, $\nabla A^*(\omegav_t) = \etav_t$ for all $t \geq 0$.
Thus, the MD update \eqref{eq:mirror-descent-update-dual-space-form} recovers the NGD update \eqref{eq:natural-gradient-update-inverse-free} exactly.

The discussion in this section so far is summarized in the lemma below.
In particular, we will not distinguish between NGD and MD in the rest of the paper.
\begin{lemma}[NGD $=$ MD]
Suppose the NGD update \eqref{eq:natural-gradient-update} and the MD update \eqref{eq:mirror-descent-update-proximal-form} start from the same variational distribution $q_0$, \ie, $\etav_0 = \nabla A^*(\omegav_0)$.
Then, we have $\etav_t = \nabla A^*(\omegav_t)$ for all $t \geq 0$.
Namely, NGD and MD produce exactly the same sequence of variational distributions.
\end{lemma}

We introduce a few definitions useful for later proofs.
Our results in \S\ref{sec:convergence} are built upon casting NGD as a special case of MD and utilizing the recent developments of stochastic MD for relatively smooth and relatively strongly convex functions \citep{birnbaum2011distributed,bauschke2017descent,lu2018relatively,hanzely2021fastest}.
\begin{definition}
Let $\Phi$ be a differentiable and strictly convex function.
A function $f$ is called $\beta$-smooth relative to $\Phi$ if
\begin{align*}
f(\av) \leq f(\bv) + \inner{\nabla f(\bv), \av - \bv} + \beta \Ds_{\Phi}(\av, \bv)
\end{align*}
holds for all $\av, \bv$ in the domain.
A function $f$ is called $\alpha$-strongly convex relative to $\Phi$ if
\begin{align*}
f(\av) \geq f(\bv) + \inner{\nabla f(\bv), \av - \bv} + \alpha \Ds_{\Phi}(\av, \bv)
\end{align*}
holds for all $\av, \bv$ in the domain.
\end{definition}
Relative smoothness and relative strongly convexity recover the usual definitions of smoothness and strong convexity when $\Phi(\cdot) = \frac12 \norm{\cdot}^2$.

\section{Stochastic Natural Gradient VI}
\label{sec:implementing-stochastic-ngvi}

This section discusses the implementation of natural gradient variational inference in the stochastic setting.
Two types of stochasticity may arise in pratice:
(a) the expected log likelihood $\Eb_{q(\zv)} \sbb{\log p(\yv \mid \zv)}$ in the ELBO \eqref{eq:elbo} does not have a closed-form for most non-conjugate likelihoods due to the intractable integral, and thus one needs to estimate it stochastically;\footnote{Though, the expected log likelihood does have a closed-form for some non-conjugate likelihoods. See \S\ref{sec:landscape} for an example.}
(b) the expected log likelihood is a finite sum over a large number of training data, and one needs to employ mini-batch stochastic optimization, \eg, \Cref{exm:linear-regression}.

Care is required when implementing the update rule \eqref{eq:natural-gradient-update-inverse-free}, or equivalently the update rule \eqref{eq:mirror-descent-update-dual-space-form}, in the stochastic setting.
Recall that the natural parameter $\etav \in \Dc$ has a domain.
In the stochastic setting, implementing the update rule \eqref{eq:natural-gradient-update-inverse-free} with stochastic gradients $\widehat\nabla \ell(\omegav_t) \approx \nabla \ell(\omega_t)$ does not necessarily guarantee that $\etav_{t+1}$ stays inside the domain $\Dc$, in which case NGD breaks down.

\subsection{A Sufficient Condition for Valid NGD Updates}
We will give a sufficient condition on the stochastic gradient $\widehat\nabla \ell(\omegav_t)$ that guarantees the natural parameter $\etav$ always stays inside the domain $\Dc$.
As shown in \S\ref{sec:exponential-family}, the KL divergence has a closed-form gradient \wrt the expectation parameter:
\begin{align*}
    \nabla_{\omegav} \Ds_\KL(q(\zv), p(\zv)) = \etav - \etav_\mathrm{p},
\end{align*}
where $\etav$ and $\etav_\mathrm{p}$ are the natural parameters of $q(\zv)$ and $p(\zv)$ respectively.
Thus, the stochasticity comes solely from stochastically estimating the expected log likelihood:
\begin{align*}
    \widehat\nabla \ell(\omegav) = - \widehat\nabla_{\omegav} \Eb_{q(\zv)} \sbb{\log p(\yv \mid \zv)} + \etav - \etav_{\mathrm{p}}.
\end{align*}
Plugging it into the NGD update \eqref{eq:natural-gradient-update-inverse-free}, we obtain
\begin{align*}
    \etav_{t + 1} = (1 - \gamma_t) \etav_t + \gamma_t \bb[\big]{\widehat\nabla_{\omegav} \Eb_{q_t(\zv)} \sbb{\log p(\yv \mid \zv)} + \etav_\mathrm{p}}.
\end{align*}
Recall that the natural parameter space $\Dc$ is an open convex set.
Hence, $\etav_{t+1}$ stays in $\Dc$ provided that
(a) $\gamma_t \in \sbb{0, 1}$ and
(b) $\widehat\nabla_{\omegav} \Eb_{q_t(\zv)} \sbb{\log p(\yv \mid \zv)} + \etav_\mathrm{p} \in \Dc$.
The first condition is satisfied if the step size $\gamma_t$ is chosen properly.
The second condition is more complicated, but can still be satisfied by carefully constructed stochastic gradient estimators.
\jrg{Example 2 is too long to be an example in italics. Maybe it should be a \textbackslash paragraph?}
\kw{fixed.}

\subsection{Common Stochastic Gradient Estimators}

This section discusses a common special case: (a) the variational family $\Qc$ is the collection of all Gaussians; (b) the prior $p(\zv)$ is a Gaussian; and (c) the likelihood $p(\yv \mid \zv)$ is log-concave in $\zv$.
In the following, we give two examples of stochastic gradients.
One example guarantees valid NGD updates while the other one does not.

For Gaussians, the only constraint on the natural parameter $\etav = (\lambdav, \Lambdav)$ is that its second component is negative definite $\Lambdav \prec 0$.
The sufficient condition for valid stochastic NGD updates in the previous section reduces to the following:
\begin{remark}
\label{rmk:gaussian-valid-ngd-condition}
Suppose that the variational and the prior are both Gaussians.
The NGD update \eqref{eq:natural-gradient-update-inverse-free} is valid for all $t \geq 0$ in the stochastic setting if
(a) the step size $\gamma_t \in [0, 1]$ and
(b) the stochastic gradient of the expected log likelihood
\begin{align*}
    \bb{\widehat\nabla_{\xiv}, \widehat\nabla_{\Xiv}} = \widehat\nabla_{\omegav} \Eb_{q(\zv)} \log p(\yv \mid \zv)
\end{align*}
has its second component negative definite $\widehat\nabla_{\Xiv} \prec 0$.
\end{remark}
\textbf{Automatic Differentiation.}
For intractable expected log likelihoods, a simple estimator for their gradients
uses the reparameterization trick \citep[][]{kingma2013auto,titsias2014doubly,rezende2014stochastic} and automatic differentiation, shown in \Cref{alg:autodiff}.
This stochastic gradient guarantees that $\widehat\nabla_{\Xiv} \Eb_q \log p(\yv \mid \zv)$ is unbiased and symmetric \citep{murray2016differentiation}, but does not guarantee $\widehat\nabla_{\Xiv}$ is negative definite.
A counterexample is given in \S\ref{sec:counter-example-autodiff-stochastic-gradient}.

\begin{algorithm2e}[t]
\caption{Auto Differentiation Stochastic Gradient}
\label{alg:autodiff}
\DontPrintSemicolon

\KwIn{$\omegav = (\xiv, \Xiv)$, the expectation parameter of $q(\zv)$}
\KwOut{$\bb[\big]{\widehat\nabla_{\xiv}, \widehat\nabla_{\Xiv}} = \widehat\nabla_{\omegav} \Eb_{q(\zv)} \sbb{\log p(\yv \mid \zv)}$}

$\bb{\muv, \Sigmav} = \bb[\big]{\xiv, \Xiv - \xiv \xiv^\top}$ 
\tcp*{conversion}

$\Cv = \texttt{cholesky}\bb{\Sigmav}$

$\uv \sim \Nc(\zero, \Iv)$

$\zv = \muv + \Cv \uv$
\tcp*{$\zv \sim \Nc(\muv, \Sigmav)$}

$\texttt{loss} = \log p(\yv \mid \zv)$
\tcp*{forward pass}

\texttt{loss.backward()}
\tcp*{compute $\widehat\nabla_{\xiv}, \widehat\nabla_{\Xiv}$}
\end{algorithm2e}

\textbf{Bonnet's and Price's Gradients.}
Consider the gradients \wrt the mean and covariance of a Gaussian $\Nc(\muv, \Sigmav)$.
By the Bonnet and Price theorems \citep{bonnet1964transformations,price1958useful,opper2009variational}, they are
\begin{align*}
    \nabla_{\muv} \Eb_{q(\zv)}\sbb{\log p(\xv \mid \zv)} & = \Eb_{q(\zv)} \sbb{\nabla_{\zv} \log p(\xv \mid \zv)}, \\
    \nabla_{\Sigmav} \Eb_{q(\zv)}\sbb{\log p(\xv \mid \zv)} & = \frac12 \Eb_{q(\zv)} \sbb{\nabla_{\zv}^2 \log p(\xv \mid \zv)}.
\end{align*}
Applying the chain rule through $\xiv = \muv$ and $\Xiv = \muv \muv^\top + \Sigmav$, and approximating the expectations with samples, we obtain a stochastic gradient $\widehat\nabla_{\omegav} \Eb_{q(\zv)}\sbb{\log p(\yv \mid \zv)}$:
\begin{align}
\label{eq:stochastic-gradient-bonnet-price}
\begin{split}
    \widehat\nabla_{\xiv} & \!=\! \frac{1}{m}\sum_{i=1}^{m}\sbb[]{\nabla_{\zv} \log p(\yv \mid \zv_i) \!-\! \nabla_{\zv}^2 \log p(\yv \mid \zv_i) \!\cdot\! \muv} \\
    \widehat\nabla_{\Xiv} & = \frac{1}{2 m} \sum_{i=1}^{m} \nabla_{\zv}^2 \log p(\yv \mid \zv_i)
\end{split}
\end{align}
where $\zv_i \sim q$ are \iid samples from the variational distribution.
While the stochastic gradient $\widehat\nabla_{\xiv}$ in \eqref{eq:stochastic-gradient-bonnet-price} coincides with the reparameterization trick, the second line is not the same as the stochastic gradient by automatic differentiation in \Cref{alg:autodiff}: $\widehat\nabla_{\Xiv}$ is negative definite for all log-concave likelihoods (concavity in $\zv$).
Hence, \eqref{eq:stochastic-gradient-bonnet-price} guarantees valid stochastic NGD updates provided that $\gamma_t \in \sbb{0, 1}$, and often appears in the natural gradient variational inference literature \citep[\eg,][]{khan2015kullback,khan2017conjugate,zhang2018noisy,lin2020handling}.

\textbf{Additional Discussion.}
The main goal of this section is to point out the sufficient condition for valid NGD updates in the stochastic setting, as well as its special case \Cref{rmk:gaussian-valid-ngd-condition}.
Those observations, though simple, are prerequisites for the convergence of stochastic NGVI in \S\ref{sec:convergence}.
Moreover, the Bonnet and Price stochastic gradients will be used in the experiments in \S\ref{sec:experiment}.

We mention a few common workarounds to take advantage of automatic differentiation, even though natively applying automatic differentiation may break down stochastic NGD.
Numerous \emph{approximate} NGD methods admit valid updates in the stochastic settings \citep{khan2018fast,osawa2019practical,lin2020handling}, with some specifically addressing the constraint on the natural parameter \citep{lin2020handling}.
An alternative is to parameterize the variational distribution with an unconstrained parameter, \eg, the mean and the covariance square root.
Refer to \citet{salimbeni2018natural} for more examples of parameterizations.
As a side effect, changing the parameterization also changes the ELBO landscape and may slow down the convergence.

\section{Convergence of Stochastic NGVI}
\label{sec:convergence}

Even though NGVI is known to converge in one step for conjugate likelihoods\jrg{cite}\kw{I actually don't have a reference for now. I feel like this is folk wisdom and I have not seen a proof yet}, it generally does not in the stochastic setting.
This section aims to establish a convergence rate of stochastic NGVI for conjugate likelihoods.
The main techniques we will use are recent developments of stochastic mirror descent for relatively smooth and strongly convex functions \citep{lu2018relatively,hanzely2021fastest}.

\begin{definition}[\citealp{hanzely2021fastest}]
Given the step sizes $\cbb{\gamma_t}_{t=0}^{\infty}$ and the iterates $\cbb{\omegav_t}_{t=0}^{\infty}$ generated by the updates \eqref{eq:mirror-descent-update-proximal-form}, we define the gradient variance at the step $t$ as
\begin{align}
\label{eq:gradient-variance}
    \frac{1}{\gamma_t} \Eb\sbb{\inner{\widehat\nabla \ell(\omegav_t) - \nabla \ell(\omegav_t), \omegav_{t + 1, *} - \omegav_{t + 1}} \mid \omegav_t},
\end{align}
where $\omegav_{t+1, *} = \argmin_{\omegav \in \Omega} \nabla \ell(\omegav_t)^\top \omegav + \frac{1}{\gamma_t} \Ds_{A^*}(\omegav, \omegav_t)$
and the conditional expectation is taken over the randomness of the stochastic gradient $\widehat \nabla \ell(\omegav_t)$.
\end{definition}
Note that the gradient variance \eqref{eq:gradient-variance} reduces to the familiar one $\Eb \norm{\widehat\nabla \ell(\omegav_t) - \nabla \ell(\omegav_t)}^2$ for gradient descent updates $\omegav_{t+1, *} \!= \omegav_t - \gamma_t \nabla \ell(\omegav_t)$ and $\omegav_{t+1} \!=\! \omegav_t - \gamma_t \widehat\nabla \ell(\omegav_t)$.
For mirror descent, however, \eqref{eq:gradient-variance} is a generalization that does not depend on a norm.
The norm-independency is crucial for our setting.
Common stochastic mirror descent analyses require the distance generating function $\Phi$ to be strongly convex \wrt a norm and then measure the gradient variance in the dual norm \citep[\eg,][]{bubeck2015convex,lan2020first,liu2023high,nguyen2023improved,fatkhullin2024taming}.
However, as shown in \S\ref{sec:log-partition-function}, the conjugate of the log-partition function $A^*$ is not strongly convex \wrt any norms, which prevents us from measuring the gradient variance with a norm.
The absence of strong convexity in the distance generating function $A^*$ may partially explain why a precise convergence rate of stochastic natural gradient variational inference is not developed over the years.
\begin{restatable}[]{lemma}{ConjugateELBOSmoothnessStrongConvexity}
\label{thm:conjugate-elbo-smoothness-strong-convexity}
For conjugate likelihoods, the negative ELBO $\ell(\omegav)$ is $1$-smooth $1$-strongly convex relative to the convex conjugate $A^*$ of the log-partition function.
\end{restatable}
The relative $1$-smoothness and $1$-strong convexity imply that the negative ELBO is a well-conditioned objective. Besides, the first-order approximation at an arbitrary $\omegav_t \in \Omega$ is exact: 
\begin{align*}
    \ell(\omegav) = \ell(\omegav_t) + \inner{\nabla \ell(\omegav_t), \omegav - \omegav_t} + \Ds_{A^*}(\omegav, \omegav_t).
\end{align*}
With the exact gradient $\nabla \ell(\omegav)$, the mirror descent update \eqref{eq:mirror-descent-update-proximal-form}, which minimizes the first-order approximation, converges in one step with the step size $\gamma_t = 1$.
However, one-step convergence is generally not possible in the stochastic setting.
Next, we present a general convergence rate that holds for all conjugate likelihoods---the prior $p(\zv)$ and the likelihood $p(\yv \mid \zv)$ are chosen such that the posterior $p(\zv \mid \yv)$ is in the same exponential family as the prior.
\begin{assumption}
\label{asm:three-assumptions-stochastic-gradient}
The stochastic gradient $\widehat\nabla \ell(\omegav_t)$
\begin{enumerate}[leftmargin=*,noitemsep,partopsep=0pt,topsep=0pt,parsep=0pt]
\item 
respects the domain: $\etav_{t+1} \in \Dc$ for all $t \geq 0$ in \eqref{eq:natural-gradient-update-inverse-free};
\item
is unbiased: $\Eb\sbb{\widehat\nabla \ell(\omegav_t) \mid \omegav_t} = \nabla \ell(\omegav_t)$;
\item
has bounded variance: \eqref{eq:gradient-variance} is bounded by $V > 0$.
\end{enumerate}
\end{assumption}
\begin{restatable}[]{theorem}{ConvergenceStochasticNaturalGradient}
\label{thm:convergence-stochastic-natural-gradient}
Suppose the likelihood $p(\yv \mid \zv)$ is conjugate and the stochastic gradient $\widehat\nabla \ell(\omegav_t)$ satisfies \Cref{asm:three-assumptions-stochastic-gradient}.
Running $T + 1$ iterations of stochastic natural gradient descent with $\gamma_t = \frac{2}{2 + t}$ generate a point $\bar\omegav_{T+1}$ that satisfies
\begin{align}
\label{eq:stochastic-ngvi-convergence-rate}
    \Eb\sbb{\ell(\bar \omegav_{T+1})} - \min_{\omegav \in \Omega} \ell(\omegav) \leq \frac{V}{T + 2},
\end{align}
where $\bar{\omegav}_{T+1} = \frac{2}{(T + 1)(T + 2)}\sum_{t=0}^{T} (t + 1) \omegav_{t+1}$.
Let $\bar{q}_{T+1}$ be the variational distribution represented by $\bar\omegav_{T+1}$.
Then, the KL divergence to the true posterior $q^*$ is bounded by
\begin{align}
    \Eb\sbb{\Ds_\KL(\bar{q}_{T+1}, q^*)} \leq \frac{V}{T + 2}.
\end{align}
\end{restatable}
\jrg{Too many remarks in a row. Can we just make this text in e.g. a ``discussion'' or ``remarks'' paragraph?}
\kw{fixed. rewrote a new paragraph that combines the remarks}

We make two observations on the rate \eqref{eq:stochastic-ngvi-convergence-rate}.
First, the rate interpolates between stochastic and deterministic settings.
In particular, zero variance $V = 0$ implies convergence in one step.
Second, the convergence rate does not depend on the distance from the initialization $q_0$ to the true posterior $q^*$.
This leads to an interesting interpretation: no matter how far away the initialization is to the true posterior, after the first iteration $\bar\omegav_1$ always goes to a sublevel set whose size only depends on the variance $V$.
Both properties are due to the step size schedule $\gamma_t = \frac{2}{2 + t}$, in particular $\gamma_0 = 1$.
In general, linearly decreasing step sizes also guarantee convergence, but may lose these two properties.

\jrg{Why is there a remark about Lemmas 3 and 4 here, which we haven't talked about yet?}
\kw{I intended to delete this remark. I wrote this remark initially to convince myself that the proof makes sense.}
It is not entirely clear if the conditions in \Cref{asm:three-assumptions-stochastic-gradient} hold in practice at all.
In particular, \Cref{asm:three-assumptions-stochastic-gradient} requires the gradient variance \eqref{eq:gradient-variance}, defined in a non-standard form, to be bounded.
The rest of this section is devoted to this question by a case study of a common conjugate variational inference problem, where we show all conditions in \Cref{asm:three-assumptions-stochastic-gradient} indeed hold in practice.
\begin{example}[Bayesian Linear Regression]
\label{exm:linear-regression}
Consider
\begin{align*}
    p(\zv) = \Nc(\zero, \Pv), \quad p(\yv \mid \Xv, \zv) = \Nc(\Xv \zv, \sigma^2 \Iv),
\end{align*}
where the prior $p(\zv)$ is a zero-mean Gaussian and the label $\yv$ has an independent Gaussian observation noise.
The negative ELBO can be written as a finite sum
\begin{align*}
\ell(q)
& = -\Eb_{q(\zv)} \sbb{\log p(\yv \mid \Xv, \zv)} + \Ds_\KL(q(\zv), p(\zv)) \\
& = -\sum_{i=1}^{n} \Eb_{q(\zv)} \log p(y_i \mid \xv_i, \zv) + \Ds_\KL(q(\zv), p(\zv)),
\end{align*}
where $\cbb{\xv_i}_{i=1}^n$ are the rows of $\Xv \in \Rb^{n \times d}$.
Without loss of generality, we assume that the variational distribution is initialized as a standard normal distribution $q_0 = \Nc(\zero, \Iv)$.
\end{example}
\jrg{More remarks... Just write text in the paper? These aren't remarks that need numbers because you reference them elsewhere. These are just paragraphs in the paper.}
\kw{fixed}

\textbf{Data Sub-Sampling Stochastic Gradient.}
Each iteration samples $m$ data points uniformly and independently:
\begin{align*}
    \xv_{i_1}, \xv_{i_2}, \cdots, \xv_{i_m}.
\end{align*}
Each index $i_k$ is independently sampled from the uniform distribution $U\sbb{n}$.
The stochastic natural gradient $\widehat\nabla\ell(\omegav)$ is
\begin{align}
\label{eq:stochastic-gradient-data-subsampling}
\nabla_{\omegav} \sbb[\bigg]{- \frac{n}{m} \sum_{k=1}^{m} \Eb_{q} \log p(y_{i_k} \mid \xv_{i_k}, \zv) \!+\! \Ds_\KL(q, p)},
\end{align}
where $p(y_{i_k} \mid \xv_{i_k}, \zv) = \Nc(\zv^\top \xv_{i_k}, \sigma^2)$ and the expectation $\Eb_{q(\zv)} \log p(y_{i_k} \mid \xv_{i_k}, \zv)$ is computed in a closed-form.
Each stochastic NGD update can be computed in $\Oc(d^2 m)$,
while the closed-form posterior of Bayesian linear regression takes $\Oc(d^2 n + d^3)$ to compute.
Approximating the posterior via stochastic NGD is more practical for large datasets.
Indeed, it is widely used in variational Gaussian processes \citep[\eg,][]{hensman2013gaussian,salimbeni2018natural} where $n$ might be too large to even fit the data into the memory.

Now we verify the conditions in \Cref{asm:three-assumptions-stochastic-gradient}.
For each $i \in \sbb{n}$, the second component $\nabla_{\Xiv}$ of the gradient
\begin{align*}
    \bb[\big]{\nabla_{\xiv}, \nabla_{\Xiv}} = \nabla_{\omegav} \Eb_q \log p(y_i \mid \xv_i, \zv)
\end{align*}
is negative definite (see \S\ref{sec:proof-stochastic-gradient-data-subsampling}).
By \Cref{rmk:gaussian-valid-ngd-condition}, the stochastic gradient \eqref{eq:stochastic-gradient-data-subsampling} indeed respects the domain $\Dc$ and results in valid NGD updates, as long as $0 \leq \gamma_t \leq 1$.
It is clearly unbiased as each data point $\xv_{i_k}$ is sampled uniformly.
Lastly, its variance is bounded:
\begin{restatable}{lemma}{GradientVarianceDataSubsampling}
\label{thm:gradient-variance-data-subsampling}
The stochastic gradient \eqref{eq:stochastic-gradient-data-subsampling} satisfies
\begin{align}
\frac{1}{\gamma_t} \Eb\sbb{\inner{\widehat\nabla\ell(\omegav_t) \!-\! \nabla \ell(\omegav_t), \omegav_{t+1, *} \!-\! \omegav_{t+1}} \mid \omegav_t} \leq V_2,
\end{align}
where
\(
V_2 = \bb{\nu s_1 +
\frac12 \nu^2 s_2 +
2 \nu^2 b \sqrt{s_1 s_2} n +
\nu^3 b^2 s_2 n^2} \frac{n^2}{\sigma^4 m}
\),
with $\nu = \max\cbb{1, \norm{\Pv}}$, $b = \max_{1 \leq i \leq n}\norm{y_i \xv_i}$, and the empirical variances
\(
s_1 = \Eb_{j \sim U\sbb{n}}\norm[\big]{y_j \xv_j - \frac1n \sum_{i=1}^{n} y_i \xv_i}^2.
\)
and
\(
s_2 = \Eb_{j \sim U\sbb{n}} \norm[\big]{\xv_j \xv_j^\top - \frac1n \sum_{i=1}^{n} \xv_i \xv_i^\top}_\F^2
\).
\end{restatable}
The constant in \Cref{thm:gradient-variance-data-subsampling} is not necessarily tight and may be improved.
Nevertheless, it serves the purpose to show that the gradient variance is bounded by a constant.

\textbf{Application to Gaussian Process Regression.}
Our result immediately applies to stochastic variational Gaussian processes (SVGP) \citep{hensman2013gaussian}, a popular large-scale Gaussian process regression model.
SVGP training minimizes the negative ELBO of the form
\begin{align*}
    -\int p(\fv \mid \uv) q(\uv) \log p(\yv \mid \fv) \diff \fv \diff \uv + \Ds_\KL(q(\uv), p(\uv)),
\end{align*}
where the variational distribution is $q(\uv)$ with the likelihood
\begin{align*}
    p(\yv \mid \fv) & = \Nc(\yv; \zero, \sigma^2 \Iv), \\
    p(\fv \mid \uv) & = \Nc(\fv; \Kv_{\fv\uv} \Kv_{\uv\uv}^{-1} \uv, \Kv_{\fv\fv} - \Kv_{\fv\uv} \Kv_{\uv\uv}^{-1} \Kv_{\uv\fv}),
\end{align*}
and the prior $p(\uv) = \Nc(\uv; \zero, \Kv_{\uv\uv})$.
Simplify the ELBO by removing terms independent of $q(\uv)$ gives
\begin{align*}
    -\!\! \int q(\uv) \log \Nc\bb{\yv; \Kv_{\fv\uv} \Kv_{\uv\uv}^{-1} \uv, \sigma^2 \Iv} \diff \uv +\! \Ds_{\KL}(q(\uv), p(\uv)).
\end{align*}
Hence, finding the optimal variational distribution $q(\uv)$ is equivalent to Bayesian linear regression in \Cref{exm:linear-regression} with $\Xv \!=\! \Kv_{\fv\uv} \Kv_{\uv\uv}^{-1}$ and the prior covariance $\Pv = \Kv_{\uv\uv}$.
Even though the optimal variational distribution $q^*$ has a closed-form, computing it exactly needs to access the entire dataset.
Besides, $q^*$ varies after every GP hyperparameter update, and it is expensive to compute $q^*$ exactly every iteration.
Thus, a popular approach is jointly minimizing the variational parameters and the hyperparameters by mini-batch stochastic optimization.
\Cref{thm:gradient-variance-data-subsampling} together with \Cref{thm:convergence-stochastic-natural-gradient} gives a convergence rate of the variational distribution in SVGP training.
The convergence rate may also find applications in some collapsed variational inference methods \citep[\eg,][]{hensman2012fast}, where NGD is applied to a subset of latent variables in an conjugate exponential family.

\section{ELBO Landscape}
\label{sec:landscape}
In the last section, we have seen that the (negative) ELBO $\ell(\omegav)$, as a function of the expectation paremters $\omegav$, has good properties when the likelihood is conjugate (see \Cref{thm:conjugate-elbo-smoothness-strong-convexity}).
\jrg{Did we? Theorem 1 doesn't list conjugacy as an assumption, and Lemmas 3 and 4 apply to the Gaussian setting, not the arbitrary conjugate setting right? If there's a general result for conjugate likelihoods, I think it's not made very clear in the last section.}
\kw{Yes, now the Theorem 1 statement is more general, applicable to conjugate likelihoods. The variance in \Cref{thm:gradient-variance-bonnet-price,thm:gradient-variance-data-subsampling}, however, only applies to Bayesian linear regression.}
These properties are crucial for the convergence analysis.
The natural question is whether the ELBO preserves these properties for non-conjugate likelihoods.

This section studies variational inference with a Gaussian prior $p(\zv)$, a Gaussian variational family $\Qc$, and a non-Gaussian (\ie, non-conjugate) likelihood $p(\yv \mid \zv)$.
Surprisingly, we show that even when the likelihood is log-concave, the ELBO $\ell(\omegav)$ is not guaranteed to be convex in the expectation parameter.
This is in sharp contrast to the mean-square-root parameterization $(\mv, \Cv)$, with $\mv$ and $\Cv$ representing the mean and the Cholesky factor respectively, used in stochastic gradient, where the ELBO is smooth and strongly convex \citep{domke2020provable}.

Below we give two examples (with details in \S\ref{sec:case-study}) where the negative ELBO $\ell(\omegav)$ is non-convex in the expectation parameter $\omegav$, even for simple log-concave likelihoods.
To show the objective is non-convex, all we need to do is to find a dataset such that the negative ELBO is non-convex.
\jrg{Split these out so that not the entire column is in italics. Maybe just make the examples paragraph headings? Again, you never refer to the numbers I think.}
\kw{Fixed. Now they are paragraphs.}

\textbf{Logistic Regression.}
Consider an $1$-dimensional Bayesian logistic regression on the dataset $\cbb{\bb{x_i, y_i}}_{i = 1}^{n}$ with $x_i \in [-1, 1]$ and $y_i \in \cbb{-1, 1}$.
The prior $p(w, b)$ on the weight $w$ and the bias $b$ is a standard Gaussian distribution.
The negative ELBO $\ell(\omegav)$ is
\begin{align*}
    \Eb_{q(w, b)} \sbb[\Bigg]{\sum_{i=1}^{n} \log\bb[\big]{1 + \exp\bb{- y_i (w x_i + b)}} } + \Ds_\KL(q, p),
\end{align*}
where $q(w, b)$ is a Gaussian variational distribution.
Restrict the expectation parameter $\omegav$ of $q$ on the convex subset 
\begin{align*}
    \cbb{\omegav \!=\! (\zero, \Xiv): \Xiv = \diag\bb{s_1, s_2}, s_1 > 0, s_2 > 0} \subseteq \Omega,
\end{align*}
where the first component of $\omegav$ is zero and the second component is a diagonal matrix.
If $\ell(\omegav)$ was convex in $\omegav$, it would be convex in $s_2$ at least.
Taking the second-order derivative with respect to $s_2$, we have
\begin{align*}
    \nabla_{s_2}^2 \ell(\omegav) = \sum_{i = 1}^{n} \Eb \sbb{\psi_i \bb{1 - \psi_i} \bb{6 \psi_i^2 - 6 \psi_i + 1}} + \frac{1}{2 s_2^2},
\end{align*}
where $\psi_i = \psi(w x_i + b)$ with $\psi$ the sigmoid function and the expectation is taken over $(w, b) \sim q_{\omegav}$.
Note that the expectation is negative in the limit:
\begin{align*}
    \lim_{s_1, s_2 \to 0} \Eb_{q(w, b)} \sbb{\psi_i \bb{1 - \psi_i} \bb{6 \psi_i^2 - 6 \psi_i + 1}} = -\frac18.
\end{align*}
In particular, there exists an absolute constant $\delta > 0$ such that when $s_1 = s_2 = \delta$ we have
\begin{align*}
    \psi_i \bb{1 - \psi_i} \bb{6 \psi_i^2 - 6 \psi_i + 1} < -\frac{1}{16}
\end{align*}
for all $1 \leq i \leq n$.
This implies $\nabla_{s_2}^2 \ell(\omegav) < 0$ when $n \geq 8 / \delta^2$ for a particular $\omegav = (\zero, \Xiv)$ with $\Xiv = \diag(\delta, \delta)$.

\textbf{Poisson Regression.}
We choose this example because of its analytical ELBO.
Bayesian Poisson regression assumes that $y \mid \xv$ follows a Poisson distribution with the expectation
\begin{align*}
    \Eb\sbb{y \mid \xv} = \exp\bb{\wv^\top \xv}.
\end{align*}
The prior $p(\wv)$ on the weight $\wv$ is a standard Gaussian distribution.
Let $\omegav = \bb{\xiv, \Xiv}$ be the expectation parameter of the Gaussian variational distribution $q$.
The expected log likelihood $- \Eb_{q(\wv)} \log p(\yv \mid \Xv, \wv)$ is
\begin{align*}
    -\yv^\top \Xv \xiv + \sum_{i = 1}^{n} \sbb[\Big]{\exp\bb[\big]{\xiv^\top \xv_i + \frac12 \xv_i^\top \bb[\big]{\Xiv - \xiv \xiv^\top} \xv_i}},
\end{align*}
which is not convex in $\xiv$.
Compute the Hessian of $\ell(\omegav)$ \wrt $\xiv$ and evaluate it on the subset of the domain
\begin{align*}
    \cbb{\omegav \!=\! (\xiv, \Xiv): \Xiv = \xiv \xiv^\top + 2 \Iv} \subseteq \Omega.
\end{align*}
Then, we obtain
\begin{align*}
    \nabla_{\xiv}^2 \ell(\omegav) \preceq \sum_{i=1}^{n} (\xv_i^\top \xiv) (\xv_i^\top \xiv - 2)\xv_i \xv_i^\top + \nabla_{\xiv}^2 A^*(\omegav).
\end{align*}
For a fixed $\omegav = (\xiv, \Xiv)$, there exists a dataset $\cbb{(\xv_i, y_i)}_{i=1}^{n}$ such that $0 < \xv_i^\top \xiv < 2$ for all $1 \leq i \leq n$.
Now, consider the Hessian $\nabla_{\xiv}^2 \ell(\omegav)$ on the scaled dataset $\cbb{(c \xv_i, y_i)}_{i=1}^{n}$ evaluated at $\frac1c \xiv$.
As $c \to \infty$, we have found a dataset such that the Hessian is negative.

Recent work demonstrates that the negative ELBO is convex with a log-concave likelihood if the variational distribution is a Gaussian distribution with the mean-square-root parameterization \citep{domke2020provable}.
However, the above two examples show that it is not the case for the expectation parameter $\omegav$. Given that the canonical implementation of NGD is equivalent to mirror descent in the expectation parameter space, NGD may implicitly optimize a non-convex objective when the likelihood is non-conjugate even for simple log-concave likelihoods.

Nonetheless, the negative ELBO does have some convenient properties for log-concave likelihoods---it is not an arbitrary non-convex objective.
\begin{restatable}{proposition}{NoSpuriousStationaryPoint}
\label{thm:no-spurious-stationary-point}
Suppose the prior and the variational family are both Gaussians.
If the likelihood $p(\yv \mid \zv)$ is log-concave in $\zv$, then the negative ELBO $\ell(\omegav)$ as a function of the expectation parameter has a unique minimizer $\omegav^*$.
In addition, if the likelihood $p(\yv \mid \zv)$ is differentiable in $\zv$, then $\omegav^*$ is the unique stationary point of $\ell(\omegav)$.
\end{restatable}
\Cref{thm:no-spurious-stationary-point} is not surprising, since there is a differentiable bijection between the expectation parameterization and the mean-square-root parameterization.
The uniqueness of the minimizer and the stationary point is derived from strongly convexity in the mean-square-root parameterization.
Thus, stochastic NGVI may still converge to the optimum with log-concave likelihoods despite the non-convexity.

We end this section by discussing some implications.
With non-conjugate likelihoods, the negative ELBO $\ell(\omegav)$ is not strongly convex nor relatively strongly convex, since it is not even convex.
Strong convexity plays a crucial role in stochastic optimization.
Without it, stochastic gradient descent has a convergence rate of $\Oc(1 / \sqrt{T})$ under standard assumptions.
This rate is improved to $\Oc(1 / T)$ for strongly convex functions.
The fact that stochastic NGVI is implicitly optimizing a non-convex objective implies that we may need to resort to new properties of the ELBO to prove its $\Oc(1 / T)$ convergence rate for non-conjugate likelihoods, if it can achieve this rate at all.
One possibility to achieve this is the Polyak-{\L}ojasiewicz inequality \citep{karimi2016linear}.

\begin{figure}[t]
\centering
\begin{subfigure}[t]{0.48\linewidth}
    \includegraphics[width=\linewidth]{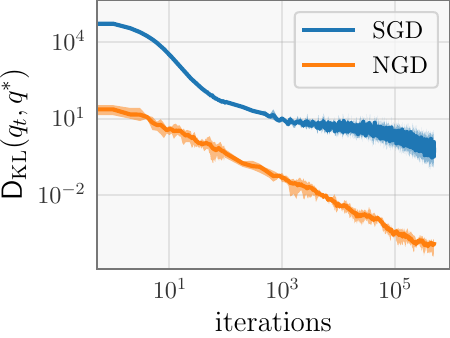}
\end{subfigure}
\begin{subfigure}[t]{0.48\linewidth}
    \includegraphics[width=\linewidth]{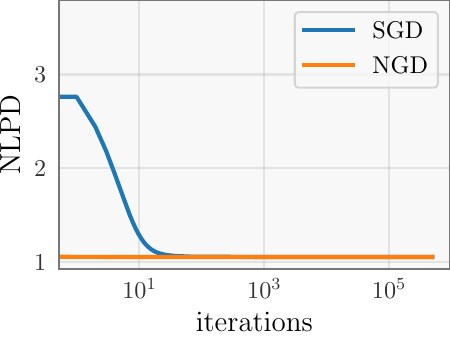}
\end{subfigure}
\caption{
Mini-batch Bayesian linear regression on the Bike dataset.
\textbf{Left:}
The KL divergence to the optimum $q^*$.
\textbf{Right:}
The training negative log predictive density.
}
\label{fig:linear-regression}
\end{figure}

\section{Numerical Simulation} 
\label{sec:experiment}

This section presents supporting numerical simulations on datasets from the UCI repository (Bike and Mushroom) and MNIST \citep{uci_datasets, mnist}

\subsection{Bayesian Linear Regression}
\Cref{fig:linear-regression} presents Bayesian linear regression on the Bike dataset ($n = 17,389$), with a standard normal prior and a noise $\sigma^2 = 1$.
The (negative) ELBO is optimized by SGD and stochastic NGD with a mini-batch size of $1000$.
SGD uses a step size schedule $\gamma_t = \frac{1}{10^{5} + t}$, a linearly decreasing schedule on the same order as \citet[][Theorem 10]{domke2023provable}.
Stochastic NGD uses a step size schedule $\gamma_t = \frac{2}{2 + t}$ predicted by \Cref{thm:convergence-stochastic-natural-gradient}.

The true posterior $q^*$ of Bayesian linear regression has a closed-form, which allows us to plot the optimality gap in the KL divergence.
In addition, we plot the negative predictive log density (NLPD) on the training set.
In the log-log scale, the KL divergences to the optimal posterior of both methods decrease at the same rate, with roughly the same slope in the figure.
This suggests that both methods have the same $\Oc(\frac1T)$ complexity, and that stochastic NGD may be only constant times faster than SGD.
Nonetheless, stochastic NGD converges very fast in the early stage.
It takes SGD thousands of iterations to catch up the progress that stochastic NGD makes in the first few iterations, implying that stochastic NGD has a much better constant factor in the big $\Oc$ notation.
Indeed, recall that the convergence rate in \Cref{thm:convergence-stochastic-natural-gradient} only depends on the stochastic gradient variance, independent of the objective's condition number and the distance from the initialization to the optimum (see \S\ref{sec:convergence}).
\jrg{Again, we have some strong guesses as to why the constants are better -- put that text somewhere and reference it here.}
\kw{added discussion.}

\subsection{Non-Conjugate Likelihoods}
\begin{figure}
\centering
\begin{subfigure}[t]{0.48\linewidth}
    \includegraphics[width=\linewidth]{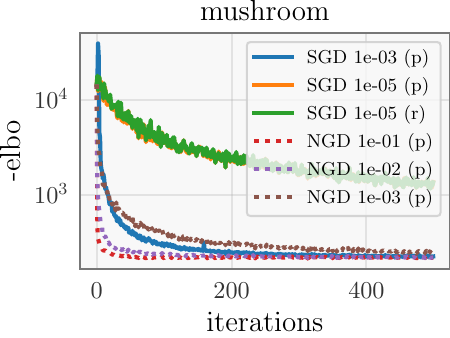}
\end{subfigure}
\begin{subfigure}[t]{0.48\linewidth}
    \includegraphics[width=\linewidth]{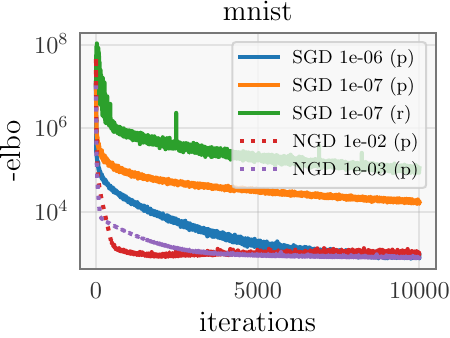}
\end{subfigure}
\caption{
Bayesian logistic regression on Mushroom and MNIST.
Labels with ``(p)'' use the stochastic gradient by the Price theorem \eqref{eq:stochastic-gradient-bonnet-price}.
Labels with ``(r)'' use the stochastic gradient by the reparamerization trick.
\vspace{-.4em}
}
\label{fig:logistic-regression}
\end{figure}

\Cref{fig:logistic-regression} shows Bayesian logistic regression on the Mushroom dataset ($n = 8124$) and MNIST (a subset of $1$ and $7$ with $n \approx 13,000$ images).
Again, stochastic NGD is faster than SGD, but the improvement is less drastic compared with conjugate likelihoods.
This is consistent with previous empirical observations \citep{salimbeni2018natural}.

Besides faster convergence, it appears that the step size of stochastic NGD is easier to tune in practice.
In most cases, the step size $\gamma \!=\! 0.1$ convergences smoothly.
Sometimes $\gamma \!=\! 0.1$ is too large such that stochastic NGD oscillates in the final stage (\Cref{fig:logistic-regression} right panel).
Simply decreasing it to $\gamma \!=\! 0.01$ leads to smooth convergence in most cases.
These observations suggest that the ELBO $\ell(\omegav)$, as a function of the expectation parameter, might have a small smoothness constant.
Indeed, the smoothness constant is $1$ for conjugate likelihoods (recall \Cref{thm:conjugate-elbo-smoothness-strong-convexity}).
For non-conjugate likelihoods in practice, we hypothesize its smoothness constant might be close to $1$ as well.

We point out a side note that the Price stochastic gradient \eqref{eq:stochastic-gradient-bonnet-price} is a high-quality gradient estimator superior to the reparameterization trick.
For instance, in the special case when the log likelihood $\log p(\yv \mid \zv)$ is a quadratic function in $\zv$, \eg, $p(\yv \mid \zv) = \Nc(\zv, \Iv)$, the Price stochastic gradient $\widehat\nabla_{\Xiv}$ is exact and has zero variance!
For general non-conjugate likelihoods, we expect the Price stochastic gradient has lower variance.
Indeed, SGD has a dramatic improvement by just switching to the Price stochastic gradient.

\section{Related Work}
Natural gradient descent was initially proposed by \citet{amari1998natural} as a learning algorithm for multi-layer perceptrons that is believed to exploit the information geometry.
Subsequently, this method has been applied to variational inference \citep[\eg,][]{honkela2004unsupervised,hensman2012fast, hoffman2013stochastic,khan2017conjugate}.
Recent new developments of natural gradient variational inference include generalization to mixtures of exponential families \citep{lin2019fast,arenz2023a}, handling the positive definite domain constraint \citep{lin2020handling}, supporting structured matrix parameterization \citep{lin2021tractable}, implementation via automatic differentiation \citep{salimbeni2018natural}, adaptations to online learning \citep{cherief2019generalization}, and generalization to Wasserstein statistical manifold \citep{chen2020optimal,li2023wasserstein}.

Outside variational inference, natural gradient descent has been applied to training (non-Bayesian) neural networks in supervised learning \citep[\eg][]{bernacchia2018exact,song2018accelerating,zhang2019fast}.
Interestingly, \citet{zhang2018noisy} establish a connection between training neural networks with noisy natural gradient and variational inference.
For a comprehensive survey of this area, we direct readers to the monograph by \citet{martens2022new}.
In particular, \citet{martens2022new} hypothesize a $\Oc(\frac1T)$ asymptotic convergence rate via an argument based on Fisher efficiency.
They also gave a non-asymptotic convergence rate of $\Oc(\frac1T)$ for stochastic preconditioned gradient descent with a fixed preconditioning matrix and a quadratic objective.

In addition, natural gradient descent has been applied to policy optimization in reinforcement learning, leading to natural policy gradient \citep{kakade2001natural}.
With the connection to mirror descent, there is a recent interest in this method that leads to a series of analyses \citep[\eg,][]{geist2019theory,shani2020adaptive,agarwal2021theory,khodadadian2021linear,xiao2022convergence,yuan2023linear}.

The natural gradient methods applied to different machine learning problems mentioned previously share a common feature: the gradient direction is preconditioned with the Fisher information matrix.
Despite being coined with the same name ``natural gradient'', we point out a subtle difference in natural gradient variational inference.
The distance generating function $A^*$ in NGVI, namely the log-partition function's conjugate, is the negative \emph{differential} entropy that is non-strongly convex and non-smooth, as shown in \S\ref{sec:log-partition-function}.
In contrast, the distance generating function in natural policy gradient is the negative \emph{Shannon} entropy, which is well-known to be strongly convex \wrt a norm \citep[\eg,][Section 4.3]{bubeck2015convex}.
As mentioned in \S\ref{sec:convergence}, this strong convexity is a key condition for mirror descent analyses in the stochastic setting.
We hope our work motivates new developments in stochastic mirror descent for non-strongly convex non-smooth distance generating functions.

\section{Conclusion}

Over the years, empirical observations suggest stochastic natural gradient descent (NGD) is faster than stochastic gradient descent for variational inference.
To understand how fast NGD converges, we prove the first $\Oc(\frac1T)$ non-asymptotic convergence rate for conjugate likelihoods.
The rate appears to be tight based on experiments, suggesting that stochastic natural gradient variational inference (NGVI) may be only constant times faster than stochastic gradient descent.
Nevertheless, the constant improvement could be dramatic in practice.
For non-conjugate likelihoods, we show that ``canonical'' stochastic NGVI implicitly optimizes a non-convex objective, which suggests that a $\Oc(\frac1T)$ rate is unlikely without discoveries of new properties of the ELBO.

\section*{Acknowledgements}
\vspace{-.2em}
The authors thank the anonymous reviewers for constructive feedbacks.
KW would like to thank Kyurae Kim for helpful discussions in the early stage of this work.
KW and JRG are supported by NSF award IIS-2145644.

\section*{Impact Statement}

This paper presents work whose goal is to advance the field of machine learning.
There might be many potential societal consequences of our work, none of which we feel must be specifically highlighted here.

\bibliography{ref}
\bibliographystyle{icml2024}

\newpage
\appendix
\onecolumn

\section{Exponential Family}
\label{sec:exponential-family}

The following useful lemma is well-known \citep[\eg,][]{nielsen2009statistical}, which hints the connection between NGD and mirror descent.
For the sake of completeness, we present a proof here, which largely follows \citet[][Lemma 2]{khan2017conjugate}.
\begin{restatable}{lemma}{BregmanEqualKL}
\label{thm:bregman-equal-kl}
Let $q(\zv)$ and $q^\prime(\zv)$ be distributions in the same exponential family.
That is, they share the same base measure $h(\zv)$, sufficient statistics $\phi(\zv)$, and the log-partition function $A(\etav)$.
Let $\omegav$ and $\omegav^\prime$ be the expectation parameters of $q(\zv)$ and $q^\prime(\zv)$ respectively.
Then, we have
\begin{align*}
    \Ds_{A^*}\bb{\omegav, \omegav^\prime} = \Ds_{A}(\etav^\prime, \etav) = \Ds_\KL\bb{q, q^\prime}.
\end{align*}
where $\Ds_{A}$ and $\Ds_{A^*}$ are the Bregman divergences associated with $A$ and $A^*$, respectively.
\end{restatable}
\begin{proof}
The first equality is a standard property of the Bregman divergence, since $\etav$ and $\omegav$ are dual to each other.
Only the second equality needs a proof:
\begin{align*}
    \Ds_\KL\bb{q, q^\prime} & = \Eb_{q(\zv)} \sbb{\log q(\zv) - \log q^\prime(\zv)} \\
    & = \Eb_{q(\zv)} \sbb{\log h(\zv) + \inner{\phi(\zv), \etav} - A(\etav)} - \Eb_{q(\zv)} \sbb{\log h(\zv) + \inner{\phi(\zv), \etav^\prime} - A(\etav^\prime)} \\
    & = \inner[\big]{\Eb_{q(\zv)} \sbb{\phi(\zv)}, \etav - \etav^\prime} - A(\etav) + A(\etav^\prime) \\
    & = A(\etav^\prime) - A(\etav) - \inner{\omegav, \etav^\prime - \etav} \\
    & = A(\etav^\prime) - A(\etav) - \inner{\nabla A(\etav), \etav^\prime - \etav} \\
    & = \Ds_A(\etav^\prime, \etav),
\end{align*}
where the second line uses the definition of the exponential family; the forth line uses the definition of the expectation parameter; the fifth line uses the duality between the natural and expectation parameters \eqref{eq:mirror-map}; and the last line uses the definition of the Bregman divergence.
\end{proof}

Let $\omegav$ and $\omegav^\prime$ be the expectation parameters of $q$ and $q^\prime$ respectively.
\Cref{thm:bregman-equal-kl} gives a simple expression for the derivative of $\Ds_\KL(q, q^\prime)$ \wrt the expectation parameter $\omegav$:
\begin{align*}
    \derivative[]{\omegav} \Ds_\KL(q, q^\prime) & = \derivative[]{\omegav} \Ds_{A^*}(\omegav, \omegav^\prime) \\
    & = \nabla A^*(\omegav) - \nabla A^*(\omegav^\prime) \\
    & = \etav - \etav^\prime,
\end{align*}
where the second line is a standard property of the Bregman divergence and the third line is because $\nabla A^*(\cdot)$ maps a expectation parameter to its corresponding natural parameter.

\subsection{Gaussian Distributions: The Log-Partition Function}
\label{sec:log-partition-function}
This section gives an explicit expression of the log-partition function $A(\etav)$ of the Gaussian distribution and its convex conjugate $A^*$.
In addition, we show that the convex conjugate $A^*$ is non-smooth and non-strongly convex.

Let $q(\zv) = \Nc(\zv; \muv, \Sigmav)$ be a $d$-dimensional Gaussian with the mean $\muv$ and the covariance $\Sigmav$.
Its density is of the form
\begin{align*}
    q(\zv; \etav) & \propto \exp\bb[\Big]{-\frac12 \bb{\zv - \muv}^\top \Sigmav^{-1} \bb{\zv - \muv} - \frac12 \log\det\bb{\Sigmav}} \\
    & = \exp\bb[\Big]{\inner{\zv, \Sigmav^{-1} \muv} + \inner{\zv \zv^\top, -\frac12 \Sigmav^{-1}}- \frac12 \muv^\top \Sigmav^{-1} \muv - \frac12 \log\det\bb{\Sigmav}}.
\end{align*}
The sufficient statistic is the map $\phi: \zv \mapsto \bb{\zv, \zv\zv^\top}$.
The natural parameter $\etav = (\lambdav, \Lambdav)$ satisfies $\lambdav = \Sigmav^{-1} \muv$ and $\Lambdav = -\frac12 \Sigmav^{-1}$.
The log-partition function $A(\cdot)$ as a function of the mean and the covariance is
\begin{align*}
    A(\muv, \Sigmav) = \frac12 \muv^\top \Sigmav^{-1} \muv + \frac12 \log\det\bb{\Sigmav}.
\end{align*}
Plug in the relation between $(\muv, \Sigmav)$ and the natural parameter $\etav = (\lambdav, \Lambdav)$.
We obtain an explicit expression of the log-partition function:
\begin{align}
\label{eq:gaussian-log-partition}
    A(\lambdav, \Lambdav) = -\frac14 \lambdav^\top \Lambdav^{-1} \lambdav - \frac12 \log\det\bb{-2 \Lambdav},
\end{align}
where $\lambdav \in \Rb^d$ and $\Lambdav \prec \zero$.
The convex conjugate $A^*$ as a function of the expectation parameter $\omegav = (\xiv, \Xiv)$ is
\begin{align}
A^*(\xiv, \Xiv) 
& =
\sup_{\lambdav \in \Rb^d, \Lambdav \preceq 0} \inner{\xiv, \lambdav} + \inner{\Xiv, \Lambdav} + \frac14 \lambdav^\top \Lambdav^{-1} \lambdav + \frac12 \log\det\bb{-2 \Lambdav}
\nonumber
\\
\label{eq:gaussian-log-partition-conjugate}
& =
-\frac12 \log \det \bb[\big]{\Xiv - \xiv \xiv^\top},
\end{align}
where the second line solves the maximization by taking the derivative and setting it equal to zero.
The constraint on the expectation parameter $(\xiv, \Xiv)$ is $\Xiv - \xiv \xiv^\top \succ 0$.

Consider the restriction of $A^*$ on the convex set $\cbb{\omegav \!=\! (\xiv, \Xiv): \xiv = \zero, \Xiv \succ 0}$.
Clearly, $A^*$ is already non-strongly convex and non-smooth in $\Xiv$: in one dimension $\frac{\diff^2}{\diff x^2} \bb{-\log x} = \frac{1}{x^2}$ is neither lower nor upper bounded.
Since the absolute value is the only norm (up to a constant) in one dimension, $A^*$ is not strongly convex \wrt any norms.
Finally, both $A$ and $A^*$ are non-smooth and non-strongly convex due to the duality between smoothness and strong convexity. 

\subsection{Gaussian Distributions: Conversion between the Natural and Expectation Parameters}
This section gives explicit expressions of $\nabla A$ and $\nabla A^*$ of Gaussian distributions.
These maps convert between the natural parameter $\etav = (\lambdav, \Lambdav)$ and the expectation parameter $\omegav = (\xiv, \Xiv)$.
Differentiating the log-partition function \eqref{eq:gaussian-log-partition} gives
\begin{align*}
\nabla_{\lambdav} A(\lambdav, \Lambdav)
& = -\frac12 \Lambdav\inv \lambdav,
\\
\nabla_{\Lambdav} A(\lambdav, \Lambdav)
& =
\frac14 \Lambdav\inv \lambdav \lambdav^\top \Lambdav\inv - \frac12 \Lambdav\inv.
\end{align*}
The gradient map exactly transforms the natural parameter to the expectation parameter, in that
\begin{align*}
\nabla_{\lambdav} A(\lambdav, \Lambdav) = \xiv,
\quad
\nabla_{\Lambdav} A(\lambdav, \Lambdav) = \Xiv.
\end{align*}
Similarly, differentiating the conjugate $A^*$ \eqref{eq:gaussian-log-partition-conjugate} gives
\begin{align*}
\nabla_{\xiv} A^*(\xiv, \Xiv)%
& = \bb[\big]{\Xiv - \xiv \xiv^\top}^{-1} \xiv,
\\
\nabla_{\Xiv} A^*(\xiv, \Xiv)%
& = -\frac12 \bb[\big]{\Xiv - \xiv \xiv^\top}^{-1},
\end{align*}
which transform the expectation parameter back to the natural parameter:
\begin{align*}
\nabla_{\xiv} A^*(\xiv, \Xiv) = \lambdav,
\quad
\nabla_{\Xiv} A^*(\xiv, \Xiv) = \Lambdav.
\end{align*}

For a Gaussian distribution $\Nc(\muv, \Sigmav)$, recall the relation between the mean/covariance and the natural parameter
\begin{align*}
    \lambdav = \Sigmav\inv \muv, \quad \Lambdav = -\frac12 \Sigmav\inv,
\end{align*}
and the relation between the mean/covariance and the expectation parameter
\begin{align*}
    \xiv = \muv, \quad\quad \Xiv = \Sigmav + \muv \muv^\top.
\end{align*}
One can verify that these relations are indeed consistent with the maps $\nabla A(\etav) = \omegav$ and $\nabla A^*(\omegav) = \etav$.

\section{Automatic Differentiation Stochastic Gradient Counterexample}
\label{sec:counter-example-autodiff-stochastic-gradient}

This section gives a counterexample where estimating the gradient of the expected log likelihood
\begin{align*}
    \widehat\nabla_{\Xiv} \Eb_{q(\zv)} \sbb{\log p(\yv \mid \zv)}
\end{align*}
using automatic differentiation shown in \Cref{alg:autodiff} does not guarantee a negative definite stochastic gradient $\widehat\nabla_{\Xiv}$.
For simplicity, we use a zero-mean Gaussian $q(\zv) = \Nc(\zero, \Sigmav)$ so that $\Xiv = \Sigmav$, and a likelihood $p(\yv \mid \zv) = \Nc(\zv, \Iv)$.
The following code produces a stochastic gradient that is not negative definite approximately $50\%$ of the time.

\begin{minipage}{\textwidth}
\begin{minted}[
    mathescape,
    linenos,
    numbersep=5pt,
    frame=lines,
]{python}
import torch
from torch.distributions import MultivariateNormal


if __name__ == "__main__":
    d = 2

    sigma = torch.eye(d).requires_grad_()
    chol = torch.linalg.cholesky(sigma)

    u = torch.randn(d)
    z = chol @ u

    dist = MultivariateNormal(z, torch.eye(d))
    y = torch.zeros(d)

    loss = dist.log_prob(y)
    loss.backward()

    print(loss.item())
    print(sigma.grad)

    # check the diagonal gradient
    # print(-0.5 * u ** 2)

    det = torch.linalg.det(sigma.grad)

    if det > 0.:
        print("not n.d.")
    else:
        print("......")
\end{minted}
\end{minipage}

\section{Stochastic Gradient Variance for Bayesian Linear Regression in \Cref{exm:linear-regression}}

In this section, we restrict ourselves to Bayesian linear regression in \Cref{exm:linear-regression}, and establish bounds on the stochastic gradient variance \eqref{eq:gradient-variance}.
Recall that the negative ELBO is a sum of the negative expected log likelihood and the KL divergence:
\begin{align*}
    \nabla \ell(\omegav) = -\nabla_{\omegav} \Eb_{q(\zv)}\sbb{\log p(\yv \mid \zv)} + \Ds_\KL(q(\zv), p(\zv)).
\end{align*}
As discussed in \S\ref{sec:exponential-family}, the KL divergence has a simple closed-form gradient available when the variational distribution $q(\zv)$ and the prior $p(\zv)$ are both in the same exponential family:
\begin{align*}
    \nabla_{\omegav} \Ds_\KL(q(\zv), p(\zv)) = \etav - \etav_\mathrm{p}.
\end{align*}
where $\etav$ and $\etav_\mathrm{p}$ are the natural parameters of $q(\zv)$ and $p(\zv)$, respectively.
Therefore, the stochasticity solely comes from estimating the expected log likelihood, and the stochastic gradient $\widehat\nabla \ell(\omegav)$ admits the form
\begin{align*}
    \widehat\nabla \ell(\omegav) = -\widehat\nabla_{\omegav} \Eb_{q(\zv)}\sbb{\log p(\yv \mid \zv)} + \nabla_{\omegav} \Ds_\KL(q(\zv), p(\zv)).
\end{align*}
For now, we assume the stochastic gradient of the expected log likelihood $(\widehat\nabla_{\xiv}, \widehat\nabla_{\Xiv}) = \widehat\nabla_{\omegav} \Eb_{q(\zv)}\sbb{\log p(\yv \mid \zv)}$ has its second component $\widehat\nabla_{\Xiv}$ negative definite, a sufficient condition for valid stochastic NGD updates.
We will show why this is the case in the upcoming section.

The next lemma shows that the natural parameter's second component $\Lambdav_t$ is bounded away from zero throughout the NGD updates, if the stochastic gradient $\widehat\nabla_{\Xiv} \Eb_{q(\zv)} \log p(\yv \mid \zv)$ is always negative definite.
\begin{lemma}
\label{thm:posterior-contraction}
For Bayesian linear regression in \Cref{exm:linear-regression}, suppose the stochastic gradient $\widehat\nabla_{\Xiv} \Eb_{q_t(\zv)} \log p(\yv \mid \Xv, \zv) \preceq 0$ and the step size $0 \leq \gamma_t \leq 1$ for all $t \geq 0$.
Then, we have $\Lambdav_t \preceq -\frac{1}{2\nu} \Iv$, or equivalently $-\frac12 \Lambdav_t\inv \preceq \nu \Iv$, throughout the NGD updates for all $t \geq 0$, where $\nu = \max\cbb{1, \norm{\Pv}} > 0$. 
\end{lemma}
\begin{proof}
We prove it by induction.
The base case $\Lambdav_0 = -\frac12 \Iv \preceq -\frac{1}{2\nu} \Iv$ satisfies the inequality trivially.
For $t \geq 1$, recall the NGD update on the natural parameter
\begin{align*}
    \etav_{t+1} = (1 - \gamma_t) \etav_t + \gamma_t \bb[\big]{\widehat\nabla_{\omegav} \Eb_{q_t(\zv)} \log p(\yv \mid \zv) + \etav_{\mathrm{p}}},
\end{align*}
where $\etav_{\mathrm{p}}$ is the natural parameter of the prior.
This yields an update on the second component of the natural parameter:
\begin{align*}
    \Lambdav_{t+1} = (1 - \gamma_t) \Lambdav_t + \gamma_t \bb[\big]{\widehat\nabla_{\Xiv} \Eb_{q_t(\zv)} \log p(\yv \mid \zv) + \Lambdav_{\mathrm{p}}}.
\end{align*}
By the assumption $\widehat\nabla_{\Xiv} \Eb_{q(\zv)} \log p(\yv \mid \zv) \preceq 0$, we have
\begin{align*}
\Lambdav_{t+1}
& \preceq (1 - \gamma_t) \Lambdav_t + \gamma_t \Lambdav_{\mathrm{p}} \\
& \preceq (1 - \gamma_t) \cdot \bb[\Big]{-\frac1{2\nu}} \cdot \Iv + \gamma_t \cdot \bb[\Big]{-\frac1{2\nu}} \cdot \Iv \\
& = -\frac1{2\nu} \Iv,
\end{align*}
where the second line uses the induction hypothesis and the fact that $\Lambdav_\mathrm{p} = -\frac12 \Pv\inv \preceq -\frac{1}{2\nu} \Iv$.
\end{proof}

The following lemma shows that the matrix inverse is a Lipschitz function in region bounded away from zero.
\begin{lemma}
\label{thm:lipschitzness-inverse}
Suppose $\Lambdav_1, \Lambdav_2 \preceq -\frac{1}{2\nu} \Iv$.
Then, we have
$\norm[]{\Lambdav_1^{-1} - \Lambdav_2^{-1}}_\F \leq 4 \nu^2 \norm{\Lambdav_1 - \Lambdav_2}_\F$.
\end{lemma}
\begin{proof}
Straightforward calculation gives a proof:
\begin{align*}
\norm[]{{\Lambdav_1^{-1} - \Lambdav_2^{-1}}}_\F & =
\norm[\big]{\Lambdav_1^{-1} - \Lambdav_2^{-1}}_\F \\
& \leq
\norm[\big]{\Lambdav_1^{-1} \bb{\Lambdav_1 - \Lambdav_2} \Lambdav_2^{-1}}_\F \\
& \leq
\norm[\big]{\Lambdav_1^{-1}} \cdot \norm{\Lambdav_1 - \Lambdav_2}_\F \cdot \norm{\Lambdav_2^{-1}} \\
& \leq
4 \nu^2 \norm{\Lambdav_1 - \Lambdav_2}_\F,
\end{align*}
where the third line uses the inequality $\norm{\Av \Bv}_\F \leq \norm{\Av} \cdot \norm{\Bv}_\F$.
\end{proof}

\textbf{Additional Notations.}
Let $\etav_t$ and $\omegav_t$ be the natural and expectation parameters of the Gaussian variational distribution $q_t$ at the step $t$.
Hence, we have $\omegav = \nabla A(\etav)$ and $\etav = \nabla A^*(\omegav)$.
Define
\begin{align*}
    \etav_{t+1, *} = \etav_t - \gamma_t \nabla \ell(\omegav_t)
\end{align*}
as the natural parameter after a NGD update from $\etav_t$ using the exact (natural) gradient $\nabla \ell(\omegav_t)$.
Recall that
\begin{align*}
    \omegav_{t+1, *} = \argmin_{\omegav \in \Omega} \inner{\nabla \ell(\omegav_t), \omegav - \omegav_t} + \frac{1}{\gamma_t} \Ds_{A^*} (\omegav, \omegav_t)
\end{align*}
is the expectation parameter after a mirror descent update from $\omegav_t$ using the exact gradient $\nabla \ell(\omegav_t)$.
Recall the relation $\omegav_{t+1, *} = \nabla A(\etav_{t+1, *})$ based on the equivalence of NGD and mirror descent.
The components of $\etav_{t+1, *}$ and $\omegav_{t+1,*}$, \ie
\begin{align*}
    \etav_{t+1, *} = (\lambdav_{t+1, *}, \Lambdav_{t+1, *}),
    \quad \quad \omegav_{t+1, *} = (\xiv_{t+1, *}, \Xiv_{t+1, *}),
\end{align*}
are marked with ``$*$'' in the subscript as well.

\subsection{Data Sub-Sampling Stochastic Gradient}
\label{sec:proof-stochastic-gradient-data-subsampling}
The stochastic gradient \eqref{eq:stochastic-gradient-data-subsampling} uses the following the estimate of the expected log likelihood 
\begin{align*}
\widehat\nabla_{\omegav} \Eb_{q(\zv)} \sbb{\log p(\yv \mid \zv)}
& =
\nabla_{\omegav} \sbb[\bigg]{\frac{n}{m} \sum_{k=1}^{m} \Eb_{q(\zv)} \log p(y_{i_k} \mid \xv_{i_k}, \zv)}
\\
& =
\nabla_{\omegav} \sbb[\bigg]{(-1) \cdot \frac{1}{\sigma^2} \frac{n}{m} \sum_{k=1}^{m} \Eb_{q(\zv)} \sbb[\Big]{\frac12 \bb{y_{i_k} - \zv^\top \xv_{i_k}}^2}}
\\
& =
\nabla_{\omegav} \sbb[\bigg]{(-1) \cdot \frac{1}{\sigma^2} \frac{n}{m} \sum_{k=1}^{m} \Eb_{q(\zv)} \sbb[\Big]{\frac12 \bb{\zv^\top \xv_{i_k}}^2 - y_{i_k} \zv^\top \xv_{i_k} + \frac12 y_{i_k}^2}}
\\
& =
-\frac{1}{\sigma^2} \frac{n}{m} \nabla_{\omegav} \sbb[\bigg]{\sum_{k=1}^{m} \bb[\Big]{\frac12 \inner[\big]{\xv_{i_k} \xv_{i_k}^\top, \Xiv} - \inner{y_{i_k} \xv_{i_k}, \xiv}}},
\end{align*}
where we note that the stochastic gradient's second component $\widehat\nabla_{\Xiv} \Eb_{q(\zv)} \sbb{\log p(\yv \mid \zv)}$ is indeed negative definite---a requirement for the NGD updates to stay inside the domain (recall \Cref{asm:three-assumptions-stochastic-gradient}).
We obtain a concrete expression for the data sub-sampling stochastic gradient $\widehat\nabla \ell(\omegav) = \bb[\big]{\widehat\nabla_{\xiv} \ell(\omegav), \widehat\nabla_{\Xiv} \ell(\omegav)}$ as follows:
\begin{align*}
    \widehat\nabla_{\xiv} \ell(\omegav_t) & = - \frac{1}{\sigma^2} \frac{n}{m} \sum_{k=1}^{m} y_{i_k} \xv_{i_k} + \lambdav - \lambdav_{\mathrm{p}}, \\
    \widehat\nabla_{\Xiv} \ell(\omegav_t) & = \frac{1}{\sigma^2} \frac{n}{m} \sum_{k=1}^{m} \frac12 \xv_{i_k} \xv_{i_k}^\top + \Lambdav - \Lambdav_{\mathrm{p}},
\end{align*}
where $\etav = \bb{\lambdav, \Lambdav}$ is the natural parameter of the variational distribution $q(\zv)$ and $\etav_\mathrm{p} = \bb{\lambdav_\mathrm{p}, \Lambdav_\mathrm{p}}$ is the natural parameter of the prior $p(\zv)$.
Meanwhile the exact gradient is
\begin{align*}
    \nabla_{\xiv} \ell(\omegav_t) & = - \frac{1}{\sigma^2} \sum_{i=1}^{n} y_i \xv_i + \lambdav - \lambdav_{\mathrm{p}}, \\
    \nabla_{\Xiv} \ell(\omegav_t) & = \frac{1}{\sigma^2} \sum_{i=1}^{n} \frac12 \xv_i \xv_i^\top + \Lambdav - \Lambdav_{\mathrm{p}}.
\end{align*}

\textbf{Roadmap.}
We give a brief overview before diving into the detailed proof of \Cref{thm:gradient-variance-data-subsampling}, which involves a large amount of (somewhat tedious) calculation.
\Cref{thm:stochastic-gradient-subsampling-first-euclidean-norm-bound,thm:stochastic-gradient-subsampling-second-euclidean-norm-bound} give bounds on the gradient variances \(
\Eb\sbb[\big]{\norm{\widehat\nabla_{\xiv} \ell(\omegav_t) - \nabla_{\xiv} \ell(\omegav_t)}^2 \mid \omegav_t}
\)
and
\(
\Eb\sbb[\big]{\norm{\widehat\nabla_{\Xiv} \ell(\omegav_t) - \nabla_{\Xiv} \ell(\omegav_t)}_\F^2 \mid \omegav_t}
\)
measured in the Euclidean norm.
These two bounds, however, are not quite enough for the convergence proof, as the desired gradient variance \eqref{eq:gradient-variance} does not depend on a specific norm.
\Cref{thm:bound-difference-expectation-parameter-first-component,thm:bound-difference-expectation-parameter-second-component}
bound $\norm{\xiv_{t+1, *} - \xiv_{t+1}}$ and $\norm{\Xiv_{t+1, *} - \Xiv_{t+1}}_\F$ with $\norm{\widehat\nabla_{\xiv} \ell(\omegav_t) - \nabla_{\xiv} \ell(\omegav_t)}$ and $\norm{\widehat\nabla_{\Xiv} \ell(\omegav_t) - \nabla_{\Xiv}\ell(\omegav_t)}_\F$.
\Cref{thm:gradient-variance-data-subsampling} utilizes \Cref{thm:bound-difference-expectation-parameter-first-component,thm:bound-difference-expectation-parameter-second-component} to reduce the gradient variance \eqref{eq:gradient-variance}, a norm-independent one, to the usual gradient variance measured in the Euclidean norm, which is readily tackled by \Cref{thm:stochastic-gradient-subsampling-first-euclidean-norm-bound} and \Cref{thm:stochastic-gradient-subsampling-second-euclidean-norm-bound}.
\begin{lemma}
\label{thm:stochastic-gradient-subsampling-first-euclidean-norm-bound}
The following inequality holds:
\begin{align*}
\Eb\sbb[\big]{\norm{\widehat\nabla_{\xiv} \ell(\omegav_t) - \nabla_{\xiv} \ell(\omegav_t)}^2 \mid \omegav_t}
=
\frac{n^2}{m} \frac{s_1}{\sigma^4}
\end{align*}
where we recall that $s_1 = \Eb_{j \sim U\sbb{n}}\norm[\big]{y_j \xv_j - \frac1n \sum_{i=1}^{n} y_i \xv_i}^2$ is the variance of $y_j \xv_j$.
\end{lemma}
\begin{proof}
Straightforward calculation gives a proof:
\begin{align*}
\Eb\sbb[\big]{\norm{\widehat\nabla_{\xiv} \ell(\omegav_t) - \nabla_{\xiv} \ell(\omegav_t)}^2 \mid \omegav_t}
& =
\frac{1}{\sigma^4} \Eb\norm[\bigg]{\frac{n}{m} \sum_{k=1}^{m} y_{i_k} \xv_{i_k} - \sum_{i=1}^{n} y_i \xv_i}^2
\\
& =
\frac{1}{\sigma^4} \frac{n^2}{m^2} \Eb\norm[\bigg]{\sum_{k=1}^{m} \bb[\Big]{y_{i_k} \xv_{i_k} - \frac1{n} \sum_{i=1}^{n} y_i \xv_i}}^2
\\
& =
\frac{1}{\sigma^4} \frac{n^2}{m} \Eb_{j \sim U\sbb{n}}\norm[\Big]{y_j \xv_j - \frac1n \sum_{i=1}^{n} y_i \xv_i}^2
\\
& =
\frac{n^2}{m} \frac{s_1}{\sigma^4}
\end{align*}
where the third line uses the fact that $i_k$'s are independently sampled from the uniform distribution $U\sbb{n}$.
\end{proof}

\begin{lemma}
\label{thm:stochastic-gradient-subsampling-second-euclidean-norm-bound}
The following inequality holds: 
\begin{align}
\Eb\sbb[\big]{\norm{\widehat\nabla_{\Xiv} \ell(\omegav_t) - \nabla_{\Xiv} \ell(\omegav_t)}_\F^2 \mid \omegav_t}
=
\frac14 \frac{n^2}{m} \frac{s_2}{\sigma^4},
\end{align}
where we recall that $s_2 = \Eb_{j \sim U\sbb{n}} \norm[\big]{\xv_j \xv_j^\top - \frac1n \sum_{i=1}^{n} \xv_i \xv_i^\top}_\F^2$ is the variance of $\xv_j \xv_j^\top$.
\end{lemma}
\begin{proof}
A straightforward calculation gives a proof:
\begin{align*}
\Eb\sbb[\big]{\norm{\widehat\nabla_{\Xiv} \ell(\omegav_t) - \nabla_{\Xiv} \ell(\omegav_t)}_\F^2 \mid \omegav_t}
& =
\Eb\norm[\Big]{\frac12 \frac{n}{m} \sum_{k=1}^{m} \xv_{i_k} \xv_{i_k}^\top - \frac12 \sum_{i=1}^{n} \xv_i \xv_i^\top}_\F^2
\\
& =
\frac14 \frac{n^2}{m^2} \Eb \norm[\Big]{\sum_{k=1}^{m} \bb[\Big]{\xv_{i_k} \xv_{i_k}^\top - \frac1n \sum_{i=1}^{n} \xv_i \xv_i^\top}}_\F^2
\\
& =
\frac14 \frac{n^2}{m} \Eb_{j \sim U\sbb{n}} \norm[\Big]{\xv_j \xv_j^\top - \frac1n \sum_{i=1}^{n} \xv_i \xv_i^\top}_\F^2
\\
& =
\frac14 \frac{n^2}{m} \frac{s_2}{\sigma^4},
\end{align*}
where the third line uses the fact that $i_k$'s are sampled independently from the uniform distribution $U\sbb{n}$.
\end{proof}

Our proof strategy is to relate the desired gradient variance \eqref{eq:gradient-variance} with the gradient variances in \Cref{thm:stochastic-gradient-subsampling-first-euclidean-norm-bound,thm:stochastic-gradient-subsampling-second-euclidean-norm-bound}.
To establish the relation, we need to show the natural parameter's first component $\lambdav_t$ and the expectation parameter's first component $\xiv_t$ are bounded throughout the NGD updates.
The trick is to observe that the natural parameter's first component $\lambdav_t$ stays in a particular region:
\begin{lemma}
Define the convex set
\(
\Cc = \cbb[\big]{\sum_{i=1}^{n}\rho_i y_i \xv_i: \rho_i \geq 0, \sum_{i=1}^{n} \rho_i \leq n}
\).
Then, we have $\lambdav_{t} \in \Cc$ and $\lambdav_{t+1, *} \in \Cc$ throughout the NGD updates for all $t \geq 0$.
\end{lemma}
\begin{proof}
We prove $\lambdav_t \in \Cc$ by induction.
The base case $t = 0$ holds as the initialization $q_0 = \Nc(\zero, \Iv)$ has $\lambdav_0 = \zero$, with coefficients $\rho_1 = \rho_2 = \cdots = \rho_n = 0$.
For $t \geq 1$, recall the update from $t$ to $t + 1$:
\begin{align*}
\lambdav_{t+1}
& = (1 - \gamma_t) \lambdav_t + \gamma_t \bb[\big]{\widehat\nabla_{\xiv} \Eb_{q_t(\zv)}\sbb{\log p(\xv \mid \zv)} + \lambdav_{\mathrm{p}}} \\
& = (1 - \gamma_t) \lambdav_t + \gamma_t \widehat\nabla_{\xiv} \Eb_{q_t(\zv)}\sbb{\log p(\xv \mid \zv)},
\end{align*}
where the second line uses $\lambdav_{\mathrm{p}} = \zero$ since the prior $p(\zv)$ is a zero-mean Gaussian.
Recall that the stochastic gradient of the expected log likelihood at the step $t$ is of the form
\begin{align*}
\widehat\nabla_{\xiv} \Eb_{q_t(\zv)}\sbb{\log p(\xv \mid \zv)} = \frac{n}{m} \sum_{k=1}^{m} y_{i_k} \xv_{i_k},
\end{align*}
where $i_k$'s are sampled independently and uniformly from $\cbb{1, 2, \cdots, n}$.
The stochastic gradient $\widehat\nabla_{\xiv} \Eb_{q_t(\zv)}\sbb{\log p(\xv \mid \zv)}$ is in the convex set $\Cc$, since the sum of its coefficients is exactly $n$.
Observe that $\lambdav_{t+1}$ is a convex combination of two points in $\Cc$, and thus stays in $\Cc$ as well.
The proof is completed by an induction.

The argument for $\lambdav_{t+1, *} \in \Cc$ follows similarly, because the exact gradient of $\nabla_{\xiv} \Eb_{q_t(\zv)} \sbb{\log p(\xv \mid \zv)} = \sum_{i=1}^{n} y_i \xv_i$ is in the convex set $\Cc$ as well.
\end{proof}
As a result, we immediately obtain a bound on the first component of the natural parameter:
\begin{corollary}
\label{thm:bound-natural-parameter-first-component}
We have $\norm{\lambdav_{t}} \leq b n$ and $\norm{\lambdav_{t, *}} \leq b n $ for all $t \geq 0$, where $b = \max_{1 \leq i \leq n} \norm{y_i \xv_i}$.
\end{corollary}
\begin{proof}
Straightforward calculation gives a proof:
\begin{align*}
\norm{\lambdav_t}
= \norm[\Big]{\sum_{i=1}^{n} \rho_i y_i \xv_i}
\leq \sum_{i=1}^{n} \rho_i \norm{y_i \xv_i}
\leq b n.
\end{align*}
The proof for $\lambdav_{t, *}$ follows the same steps.
\end{proof}

As a result, we also obtain a bound on the first component of the expectation parameter:
\begin{corollary}
\label{thm:bound-expectation-parameter-first-component}
We have $\norm{\xiv_t} \leq \nu b n$ and $\norm{\xiv_{t, *}} \leq \nu b n$ for all $t \geq 0$, where $b = \max_{1 \leq i \leq n} \norm{y_i \xv_i}$.
\end{corollary}
\begin{proof}
Recall the relation between the natural and expectation parameters: $\xiv_t = -\frac12 \Lambdav_t^{-1} \lambdav_t$.
Recall that $0 \preceq -\frac12 \Lambdav_t\inv \preceq \nu \Iv$ by \Cref{thm:posterior-contraction}.
Thus, we have $\norm{\xiv_t} \leq \nu \norm{\lambdav_t} \leq \nu b n$.
\end{proof}

\begin{lemma}
\label{thm:bound-difference-expectation-parameter-first-component}
We have
\(
\norm{\xiv_{t+1, *} - \xiv_{t+1}}
\leq
\gamma_t \nu \norm{\widehat\nabla_{\xiv} \ell(\omegav_t) - \nabla_{\xiv} \ell(\omegav_t)}
+ 2 \gamma_t \nu^2 b n \norm[\big]{\widehat\nabla_{\Xiv} \ell(\omegav_t) - \nabla_{\Xiv} \ell(\omegav_t)}_\F
\).
\end{lemma}
\begin{proof}
Straightforward calculation gives
\begin{align*}
\norm{\xiv_{t+1, *} - \xiv_{t+1}}
& = \frac12 \norm{\Lambdav_{t+1}^{-1} \lambdav_{t+1} - \Lambdav_{t+1, *}^{-1} \lambdav_{t+1, *}} \\
& = \frac12 \norm{\Lambdav_{t+1}^{-1} \lambdav_{t+1} - \Lambdav_{t+1}^{-1} \lambdav_{t+1, *} + \Lambdav_{t+1}^{-1} \lambdav_{t+1, *} - \Lambdav_{t+1, *}^{-1} \lambdav_{t+1, *}} \\
& \leq
\frac12 \norm{\Lambdav_{t+1}^{-1} \bb{\lambdav_{t+1} - \lambdav_{t+1, *}}} +
\frac12 \norm{\bb{\Lambdav_{t+1}^{-1} - \Lambdav_{t+1, *}^{-1}} \lambdav_{t+1, *}},
\end{align*}
where the first line uses the relation between the natural and expectation parameters.
We cope with the two terms separately.

For the first term, we have
\begin{align*}
\frac12 \norm{\Lambdav_{t+1}^{-1} \bb{\lambdav_{t+1} - \lambdav_{t+1, *}}}
\leq
\nu \norm{\lambdav_{t+1} - \lambdav_{t+1, *}}
=
\gamma_t \nu \norm{\widehat\nabla_{\xiv} \ell(\omegav_t) - \nabla_{\xiv} \ell(\omegav_t)},
\end{align*}
where the first inequality uses $-\frac12 \Lambdav_{t+1}\inv \preceq \nu \Iv$ by \Cref{thm:posterior-contraction}; the second equality uses the definition of the NGD update.

For the second term, we have
\begin{align*}
\frac12 \norm{\bb{\Lambdav_{t+1}\inv - \Lambdav_{t+1, *}\inv} \lambdav_{t+1, *}}
& \leq \frac12 \norm{\Lambdav_{t+1}\inv - \Lambdav_{t+1, *}\inv}_\F \cdot \norm{\lambdav_{t+1, *}} \\
& \leq 2 \nu^2 \cdot \norm{\Lambdav_{t+1} - \Lambdav_{t+1, *}}_\F \cdot \norm{\lambdav_{t+1, *}} \\
& = 2 \gamma_t \nu^2 \norm[\big]{\widehat\nabla_{\Xiv} \ell(\omegav_t) - \nabla_{\Xiv} \ell(\omegav_t)}_\F \cdot \norm{\lambdav_{t+1, *}}
\\
& \leq 2 \gamma_t \nu^2 b n \norm[\big]{\widehat\nabla_{\Xiv} \ell(\omegav_t) - \nabla_{\Xiv} \ell(\omegav_t)}_\F,
\end{align*}
where the second line uses the Lipschitz condition in \Cref{thm:lipschitzness-inverse}; the third line uses the the definition of the NGD update; the last line uses \Cref{thm:bound-natural-parameter-first-component}.
Summing the two bounds completes the proof.
\end{proof}

\begin{lemma}
\label{thm:bound-difference-expectation-parameter-second-component}
We have
\[
\norm{\Xiv_{t+1, *} - \Xiv_{t+1}}_{\F}
\leq
2 \gamma_t \nu^2 b n \norm[\big]{\widehat\nabla_{\xiv} \ell(\omegav_t) - \nabla_{\xiv} \ell(\omegav_t)} +
\bb{2 \gamma_t \nu^2 + 4 \gamma_t \nu^3 b^2 n^2} \norm[\big]{\widehat\nabla_{\Xiv} \ell(\omegav_t) - \nabla_{\Xiv} \ell(\omegav_t)}_\F.
\]
\end{lemma}
\begin{proof}
Expanding the norm, we have
\begin{align*}
\norm{\Xiv_{t+1, *} - \Xiv_{t+1}}_{\F}
& = \norm[\Big]{-\frac12\bb[\big]{\Lambdav_{t+1, *}^{-1} - \Lambdav_{t+1}^{-1}} + \bb[\big]{\xiv_{t+1, *} \xiv_{t+1, *}^\top - \xiv_{t+1} \xiv_{t+1}^\top}}_\F \\
& \leq \frac12 \norm[\Big]{\Lambdav_{t+1, *}\inv - \Lambdav_{t+1}\inv}_\F + \norm[\big]{\xiv_{t+1, *} \xiv_{t+1, *}^\top - \xiv_{t+1} \xiv_{t+1}^\top}_\F,
\end{align*}
where the first line uses the relation between natural and expectation parameters.

For the first term, we have
\begin{align*}
\frac12 \norm[\big]{\Lambdav_{t+1, *}\inv - \Lambdav_{t+1}\inv}_\F
& \leq 2 \nu^2 \norm[\big]{\Lambdav_{t+1, *} - \Lambdav_{t+1}}_\F \\
& \leq 2 \gamma_t \nu^2 \norm[\big]{\widehat\nabla_{\Xiv} \ell(\omegav_t) - \nabla_{\Xiv} \ell(\omegav_t)}_\F,
\end{align*}
where the first line uses \Cref{thm:lipschitzness-inverse}; the second line uses the definition of the NGD update.

For the second term, we have
\begin{align*}
\norm[\big]{\xiv_{t+1, *} \xiv_{t+1, *}^\top - \xiv_{t+1} \xiv_{t+1}^\top}_\F
& =
\norm[\big]{\xiv_{t+1, *} \xiv_{t+1, *}^\top - \xiv_{t+1, *} \xiv_{t+1}^\top + \xiv_{t+1, *} \xiv_{t+1}^\top - \xiv_{t+1} \xiv_{t+1}^\top}_\F
\\
& \leq 
\norm[\big]{\xiv_{t+1, *} \bb{\xiv_{t+1, *} - \xiv_{t+1}}^\top}_\F +
\norm[\big]{\bb{\xiv_{t+1, *} - \xiv_{t+1}} \xiv_{t+1}^\top}_\F \\
& =
\norm{\xiv_{t+1, *}} \cdot \norm{\xiv_{t+1, *} - \xiv_{t+1}} +
\norm{\xiv_{t+1, *} - \xiv_{t+1}} \cdot \norm{\xiv_{t+1}}
\\
& \leq
2 \max\cbb{\norm{\xiv_{t+1, *}}, \norm{\xiv_{t+1}}} \cdot \norm{\xiv_{t+1, *} - \xiv_{t+1}}
\\
& \leq
2 \nu b n \norm{\xiv_{t+1,*} - \xiv_{t+1}} \\
& \leq
2 \gamma_t \nu^2 b n \norm[\big]{\widehat\nabla_{\xiv} \ell(\omegav_t) - \nabla_{\xiv} \ell(\omegav_t)} +
4 \gamma_t \nu^3 b^2 n^2 \norm[\big]{\widehat\nabla_{\Xiv} \ell(\omegav_t) - \nabla_{\Xiv} \ell(\omegav_t)}_\F
\end{align*}
where the third line is because $\norm{\av \bv^\top}_\F = \norm{\av} \cdot \norm{\bv}$; the fifth line uses \Cref{thm:bound-expectation-parameter-first-component}; the last line uses \Cref{thm:bound-difference-expectation-parameter-first-component}.
Summing the above two parts finishes the proof.
\end{proof}

Now we are ready to prove the main results of this section, the variance bound of data sub-sampling stochastic gradient.
\GradientVarianceDataSubsampling*
\begin{proof}
Expanding the inner product inside the expectation, we need to bound the expectation of
\begin{align*}
\inner[\big]{\widehat\nabla_{\xiv} \ell(\omegav_t) - \nabla_{\xiv} \ell(\omegav_t), \xiv_{t+1, *} - \xiv_{t+1}} +
\inner[\big]{\widehat\nabla_{\Xiv} \ell(\omegav_t) - \nabla_{\Xiv} \ell(\omegav_t), \Xiv_{t+1, *} - \Xiv_{t+1}}.
\end{align*}
For the first term
\(
\inner[\big]{\widehat\nabla_{\xiv} \ell(\omegav_t) - \nabla_{\xiv} \ell(\omegav_t), \xiv_{t+1, *} - \xiv_{t+1}}
\),
applying the Cauchy-Schwarz inequality and \Cref{thm:bound-difference-expectation-parameter-first-component} yields
\begin{align}
\label{eq:first-bound-gradient-inner-product}
\begin{split}
\inner[\big]{\widehat\nabla_{\xiv} \ell(\omegav_t) & - \nabla_{\xiv} \ell(\omegav_t), \xiv_{t+1, *} - \xiv_{t+1}}
\leq
\norm{\widehat\nabla_{\xiv} \ell(\omegav_t) - \nabla_{\xiv} \ell(\omegav_t)} \cdot \norm{\xiv_{t+1, *} - \xiv_{t + 1}}
\\
& \leq
\gamma_t \nu \norm{\widehat\nabla_{\xiv} \ell(\omegav_t) - \nabla_{\xiv} \ell(\omegav_t)}^2
+
2 \gamma_t \nu^2 b n \norm{\widehat\nabla_{\xiv} \ell(\omegav_t) - \nabla_{\xiv} \ell(\omegav_t)} \cdot \norm{\widehat\nabla_{\Xiv} \ell(\omegav_t) - \nabla_{\Xiv} \ell(\omegav_t)}_\F.
\end{split}
\end{align}
For the second term 
\(
\inner[\big]{\widehat\nabla_{\Xiv} \ell(\omegav_t) - \nabla_{\Xiv} \ell(\omegav_t), \Xiv_{t+1, *} - \Xiv_{t+1}}
\),
applying the Cauchy-Schwarz inequality and \Cref{thm:bound-difference-expectation-parameter-second-component} gives
\begin{align}
\label{eq:second-bound-gradient-inner-product}
\begin{split}
\inner[\big]{& \widehat\nabla_{\Xiv} \ell(\omegav_t) - \nabla_{\Xiv} \ell(\omegav_t), \Xiv_{t+1, *} - \Xiv_{t+1}}
\leq
\norm[\big]{\widehat\nabla_{\Xiv} \ell(\omegav_t) - \nabla_{\Xiv} \ell(\omegav_t)}_\F \cdot \norm{\Xiv_{t+1, *} - \Xiv_{t+1}}_\F
\\
& \leq
2 \gamma_t \nu^2 b n \norm[\big]{\widehat\nabla_{\xiv} \ell(\omegav_t) \!-\! \nabla_{\xiv} \ell(\omegav_t)} \cdot \norm[\big]{\widehat\nabla_{\Xiv} \ell(\omegav_t) - \nabla_{\Xiv} \ell(\omegav_t)}_\F +
\bb{2 \gamma_t \nu^2 + 4 \gamma_t \nu^3 b^2 n^2} \norm[\big]{\widehat\nabla_{\Xiv} \ell(\omegav_t) - \nabla_{\Xiv} \ell(\omegav_t)}_\F^2
\end{split}
\end{align}
Summing \eqref{eq:first-bound-gradient-inner-product} and \eqref{eq:second-bound-gradient-inner-product}, and then applying the inequality
\[
\Eb\sbb[\Big]{\norm[\big]{\widehat\nabla_{\xiv} \ell(\omegav_t) - \nabla_{\xiv} \ell(\omegav_t)} \cdot \norm[\big]{\widehat\nabla_{\Xiv} \ell(\omegav_t) - \nabla_{\Xiv} \ell(\omegav_t)}_\F}
\leq
\sqrt{
\Eb\sbb[\Big]{\norm[\big]{\widehat\nabla_{\xiv} \ell(\omegav_t) - \nabla_{\xiv} \ell(\omegav_t)}^2}
\Eb\sbb[\Big]{\norm[\big]{\widehat\nabla_{\Xiv} \ell(\omegav_t) - \nabla_{\Xiv} \ell(\omegav_t)}_\F^2}
},
\]
where the expectations are conditioned on $\omegav_t$, we obtain a bound on
as follows:
\begin{align*}
\Eb\sbb{\inner{\widehat\nabla\ell(\omegav_t) - \nabla \ell(\omegav_t), & \omegav_{t+1, *} \!-\! \omegav_{t+1}} \mid \omegav_t}
\leq
\gamma_t \nu \Eb\sbb[\big]{\norm{\widehat\nabla_{\xiv} \ell(\omegav_t) - \nabla_{\xiv} \ell(\omegav_t)}^2 \mid \omegav_t}
\\
& +
\bb{2 \gamma_t \nu^2 + 4 \gamma_t \nu^3 b^2 n^2}
\Eb\sbb[\Big]{\norm[\big]{\widehat\nabla_{\Xiv} \ell(\omegav_t) - \nabla_{\Xiv} \ell(\omegav_t)}_\F^2 \mid \omegav_t}
\\
& +
4 \gamma_t \nu^2 b n
\sqrt{
\Eb\sbb[\Big]{\norm[\big]{\widehat\nabla_{\xiv} \ell(\omegav_t) - \nabla_{\xiv} \ell(\omegav_t)}^2 \mid \omegav_t}
\Eb\sbb[\Big]{\norm[\big]{\widehat\nabla_{\Xiv} \ell(\omegav_t) - \nabla_{\Xiv} \ell(\omegav_t)}_\F^2 \mid \omegav_t}
}
\\
& \leq
\gamma_t \nu \cdot \frac{n^2}{m} \frac{s_1}{\sigma^4} + 
\bb{2 \gamma_t \nu^2 + 4 \gamma_t \nu^3 b^2 n^2} \cdot \frac14 \frac{n^2}{m} \frac{s_2}{\sigma^4} +
4 \gamma_t \nu^2 b n \cdot \frac12 \frac{n^2}{m} \frac{\sqrt{s_1 s_2}}{\sigma^4}
\\
& =
\gamma_t \nu \frac{n^2}{m} \frac{s_1}{\sigma^4} +
+ \frac12 \gamma_t \nu^2 \frac{n^2}{m} \frac{s_2}{\sigma^4}
+ 2 \gamma_t \nu^2 b \frac{n^3}{m} \frac{\sqrt{s_1 s_2}}{\sigma^4}
+ \gamma_t \nu^3 b^2 \frac{n^4}{m} \frac{s_2}{\sigma^4}
\\
& =
\frac{1}{\sigma^4} \gamma_t \bb{\nu s_1 +
+ \frac12 \nu^2 s_2
+ 2 \nu^2 b \sqrt{s_1 s_2} n 
+ \nu^3 b^2 s_2 n^2} \frac{n^2}{m}
\end{align*}
where the second equality is due to \Cref{thm:stochastic-gradient-subsampling-first-euclidean-norm-bound} and \Cref{thm:stochastic-gradient-subsampling-second-euclidean-norm-bound}.
Dividing both sides by $\gamma_t$ completes the proof.
\end{proof}

\section{Proof of the Main Theorem}

\ConjugateELBOSmoothnessStrongConvexity*
\begin{proof}
Let $q^*(\zv) = p(\zv \mid \yv)$ be the posterior.
By the definition of the negative ELBO, we have $\ell(\omegav) = \Ds_{\KL}(q, q^*) + C$, where $q \in \Qc$ is the variational distribution inside an exponential family $\Qc$ parameterized by the expectation parameter $\omegav$ and $C = p(\yv)$ is a constant (log evidence) that does not depend on $q$ and $\omegav$.

Thanks to conjugacy, the posterior $q^*$ is of the same form as $q$.
By \Cref{thm:bregman-equal-kl}, $\ell(\omegav) \propto \Ds_\KL(q, q^*) = \Ds_{A^*}(\omegav, \omegav^*)$.
Observe that the Bregman divergence $\Ds_{A^*}(\omegav, \omegav^*)$ is trivially $1$-smooth and $1$-strongly convex in $\omegav$ relative to $A^*$.
\end{proof}

Below we present the main theorem, which adapts the results by \citet{hanzely2021fastest} to stochastic natural gradient variational inference.
\ConvergenceStochasticNaturalGradient*
\begin{proof}
By the descent lemma of \citet[][Lemma 5.2]{hanzely2021fastest}, we have
\begin{align*}
    \Eb\sbb{\ell(\omegav_{t+1})} - \ell(\omegav^*) \leq \bb[\Big]{\frac{1}{\gamma_t} - 1} \Ds_{A^*}(\omegav^*, \omegav_t) - \frac{1}{\gamma_t} \Eb\sbb{\Ds_{A^*}(\omegav^*, \omegav_{t+1})} + \gamma_t V.
\end{align*}
Plugging in $\gamma_t = \frac{2}{2 + t}$, we obtain
\begin{align*}
    \Eb\sbb{\ell(\omegav_{t+1})} - \ell(\omegav^*) \leq \frac12 t \cdot \Ds_{A^*}(\omegav^*, \omegav_t) - \frac12 \bb[\big]{t + 2} \Eb\sbb{\Ds_{A^*}(\omegav^*, \omegav_{t+1})} + \gamma_t V.
\end{align*}
Multiply the inequality by $t + 1$ and sum from $0$ to $T$.
Then we have
\begin{align*}
\sum_{t=0}^{T} (t + 1) \bb{\ell(\omegav_{t+1}) - \ell(\omegav^*)} \leq \frac12 V \sum_{t=0}^{T} \frac{t + 1}{t + 2} \leq \frac12 V (T + 1).
\end{align*}
Dividing both sides by $\sum_{t=0}^{T} (t + 1) = \frac12 (T + 1) (T + 2)$, and use the convexity of $f$, we obtain
\begin{align*}
    \ell(\bar\omegav_{T+1}) \leq \frac{V}{T + 2}
\end{align*}
To get the convergence rate in terms of the KL divergence, notice that \begin{align*}
    \ell(\bar\omegav_{t+1}) - \ell(\omegav^*) & = \nabla \ell(\omegav^*)^\top (\omegav_{T+1} - \omegav^*) + \Ds_{A^*}(\bar\omegav_{T+1}, \omegav^*) \\
    & = \Ds_{A^*}(\bar\omegav_{T+1}, \omegav^*) \\
    & = \Ds_\KL(\bar{q}_{T+1}, q^*),
\end{align*}
where the first line is due to $1$-smoothness and $1$-strong convexity relative to $A^*$; the second line is because the optimal parameter $\omegav^*$ has zero gradient; the third line is due to \Cref{thm:bregman-equal-kl}.
\end{proof}

\section{Missing Proofs in \S\ref{sec:landscape}}

\NoSpuriousStationaryPoint*
\begin{proof}
Consider the set
\begin{align*}
    \Theta = \{\thetav = \bb[\big]{\muv, \Cv}: \muv \in \Rb^d, \Cv \in \Sb_{++}^d\}
\end{align*}
which parameterizes all (non-degenerate) Gaussian distributions.
Define $f(\thetav) = \bb[\big]{\muv, \Cv\Cv^\top + \muv\muv^\top}$.
Namely, $f$ maps $\thetav$ to the expectation parameter space $\Omega$.
Thanks to the uniqueness of matrix square root, $f$ is a bijection.

Since $\ell^{(\mathrm{mr})}$ is strongly convex in $\thetav$, it has a unique minimizer $\thetav^* \in \Theta$.
Define $\omegav^* = f(\thetav^*)$.
It is clear that $\omegav^* \in \Omega$ is the unique minimizer of $\ell^{(\mathrm{e})}$.

Consider the identity
\begin{align}
\label{eq:elbo-equal-two-parameterization}
    \ell^{(\mathrm{mr})} (\thetav) = \ell^{(\mathrm{e})}(f(\thetav)).
\end{align}
Taking the derivative of \eqref{eq:elbo-equal-two-parameterization} on both sides, we have
\begin{align*}
    \pderivative{\muv} \ell^{(\mathrm{mr})}(\thetav) & = \pderivative[\ell^{(\mathrm{e})}]{\xiv} \bigg\rvert_{\omegav = f(\thetav)} + 2 \cdot \pderivative[\ell^{(\mathrm{e})}]{\Xiv} \bigg\rvert_{\omegav = f(\thetav)} \cdot \muv \\
    \pderivative{\Cv} \ell^{(\mathrm{mr})}(\thetav) & = 2 \cdot \pderivative[\ell^{(\mathrm{e})}]{\Xiv} \bigg\rvert_{\omegav = f(\thetav)} \cdot \Cv,
\end{align*}
It easy to see that $\nabla \ell^{(\mathrm{mr})} (\thetav) = \zero$ iff $\nabla \ell^{(\mathrm{e})} (f(\thetav)) = \zero$.
Namely, $f$ maps stationary points to stationary points.
Since there is only one stationary point in $\Theta$ due strong convexity, there is only one stationary point in $\Omega$ as well.
\end{proof}

\label{sec:case-study}

\subsection{Bayesian Logistic Regression}
We give a more detailed description of the non-convexity of Bayesian logistic regression.
Recall that we focus on the restriction of $\ell(\omegav)$ on the convex subset
\begin{align*}
    \cbb{\omegav = (\zero, \Xiv): \Xiv = \diag\bb{s_1, s_2}, s_1 > 0, s_2 > 0} \subseteq \Omega.
\end{align*}
Observe that $w x_i + b$ follows a Gaussian distribution $\Nc(0, x_i^2 s_1 + s_2)$.
Therefore, we can use the Price theorem to take the derivative \wrt $s_2$.
Taking the first-order derivative of $\ell(\omegav)$ \wrt $s_2$, we have
\begin{align*}
    \frac{\partial}{\partial s_2} \ell(\omegav) = \sum_{i = 1}^{n} \Eb_{q(w, b)} \sbb{\psi_i \bb{1 - \psi_i}} + \frac12 - \frac{1}{2 s_2},
\end{align*}
where we use $\psi_i$ to denote $\psi(w x_i + b)$ and $\psi$ is the sigmoid function.
Using the Price theorem again to take the second-order derivative of $\ell(\omegav)$ \wrt $s_2$, we have
\begin{align*}
    \frac{\partial^2}{\partial s_2^2} \ell(\omegav) = \sum_{i = 1}^{n} \Eb_{q(w, b)}\sbb{\psi_i \bb{1 - \psi_i} \bb{6 \psi_i^2 - 6 \psi_i + 1}} + \frac{1}{2 s_2^2},
\end{align*}
Note that $\pderivative[^2]{s_2^2} \ell(\omegav)$ is continuous \wrt $s_1$ and $s_2$.
Moreover, we have
\begin{align*}
    \lim_{s_1 \to 0, s_2 \to 0} \Eb\sbb{\psi_i\bb{1 - \psi_i} \bb{6 \psi_i^2 - 6 \psi_i + 1}} = -\frac18.
\end{align*}
Therefore, there exists a small positive constant $\delta > 0$, such that $s_1 = s_2 = \delta$ and
\begin{align*}
    \Eb_{w \sim \Nc\bb{0, s_1}, b \sim \Nc\bb{0, s_2}} \sbb{\psi_i \bb{1 - \psi_i} \bb{6 \psi_i^2 - 6 \psi_i + 1}} < -\frac{1}{16}.
\end{align*}
Crucially, $\delta$ is an absolute constant that does not depend on $i$.
Because all $-1 \leq x_i \leq 1$ are bounded, the distribution $w x_i + b \sim \Nc(0, x_i^2 s_1 + s_2)$ will shrink to zero as long as $s_1 + s_2 \to 0$, regardless of the index $i$.
This implies that when $s_1 = s_2 = \delta$, we have
\begin{align*}
    \sum_{i = 1}^{n} \Eb_{w \sim \Nc\bb{0, s_1}, b \sim \Nc\bb{0, s_2}} \sbb{\psi_i \bb{1 - \psi_i} \bb{6 \psi_i^2 - 6 \psi_i + 1}}
    < -\frac{1}{16} n.
\end{align*}
Therefore, when $s_1 = s_2 = \delta$ and $n \geq \frac{8}{\delta^2}$, the second order derivative is negative
\begin{align*}
    \frac{\partial^2}{\partial s_2^2} \ell(\omegav) < -\frac{1}{16} n + \frac{1}{2 \delta^2} < 0,
\end{align*}
which implies that the objective is non-convex in the expectation parameter.

\subsection{Bayesian Poisson Regression}
\label{sec:bayesian-poisson-regression}
Bayesian Poisson regression assumes that $y \mid \xv$ follows a Poisson distribution with the expectation 
\begin{align*}
    \Eb\sbb{y \mid \xv} = \exp\bb{\wv^\top \xv},
\end{align*}
which gives the log likelihood
\begin{align*}
    \log p\bb{y \mid \xv, \wv} = - \log y! + y \wv^\top \xv - \exp\bb{\wv^\top \xv}.
\end{align*}
We impose a Gaussian prior $p(\wv) = \Nc(\zero, \Iv)$ and approximate the posterior $p(\wv \mid \yv)$ using a Gaussian variational distribution $q(\wv)$.
A nice property of the Bayesian Poisson regression is that its ELBO has a closed-form expression
\begin{align*}
    \ell(\omegav) & = \sum_{i = 1}^{n} \Eb_{q(\wv)} \sbb{-y_i\wv^\top \xv_i + \exp\bb{\wv^\top \xv_i}} + \Ds_{\mathrm{KL}}\bb[\big]{q, p} \\
    & = \sum_{i = 1}^{n} \sbb[\Big]{-y_i \xiv^\top \xv_i + \exp\bb[\Big]{\xiv^\top \xv_i + \frac12 \xv_i^\top \bb[\big]{\Xiv - \xiv \xiv^\top} \xv_i}} + \Ds_{A^*}(\omegav, \omegav_0).
\end{align*}
The Hessian $\nabla_{\xiv}^2 \ell(\omegav)$ is
\begin{align*}
    \sum_{i=1}^{n} \exp\bb[\Big]{\xiv^\top \xv_i + \frac12 \xv_i^\top \bb{\Xiv - \xiv \xiv^\top} \xv_i} \bb[\big]{-1 + \bb[\big]{1 - \xv_i^\top \xiv}^2} \xv_i \xv_i^\top + \nabla_{\xiv}^2 A^*(\omegav).
\end{align*}
Evaluating the Hessian on the subset of the domain
\begin{align*}
    \cbb{\omegav = (\xiv, \Xiv) \in \Omega: \Xiv = \xiv \xiv^\top + 2\Iv},
\end{align*}
we obtain the following
\begin{align*}
    \sum_{i=1}^{n} \exp\bb[\big]{\xv_i^\top \xiv + \xv_i^\top \xv_i} \xv_i^\top \xiv \bb{\xv_i^\top \xiv - 2} \xv_i \xv_i^\top + \nabla_{\xiv}^2 A^*(\omegav).
\end{align*}
With $0 < \xv_i^\top \xiv < 2$ for all $i$, which can be satisfied by constructing the dataset properly, and using $\exp(\xv_i^\top \xiv + \xv_i^\top \xv_i) > 1$, we can drop the exponential term.
The rest of the argument follows the main paper.

\section{Experimental Details}

In all experiments, SGD uses the $(\mv, \Cv)$ parameterization, where $\mv$ is the Gaussian mean and $\Cv$ is the Cholesky factor of the Gaussian covariance.
We parameterize $\Cv$ as a lower triangular matrix with strictly positive diagonal entries.
For SGD, we clamp the diagonal entries of $\Cv$ to make sure they are no smaller than $10^{-10}$.
This is effectively a projection step.

\subsection{Bayesian Linear Regression}
This is a Bayesian linear regression problem exactly the same as \Cref{exm:linear-regression} with a standard Gaussian prior.
Note that the expected log likelihood $\Eb_{q(\zv)} \log p(\yv \mid \Xv, \zv)$ is integrated in a closed-form.
The only stochasticity comes from the mini-batch data sub-sampling.
\citet[][Theorem 7 and Theorem 10]{domke2023provable} have proved convergence for stochastic proximal (projected) gradient descent with a step size schedule $\gamma_t = \min\cbb[\big]{\frac{\mu}{a}, \frac{1}{\mu} \frac{2t + 1}{(t + 1)^2}}$.
It is not easy to come up with a tight estimate of the constant $a$.
Therefore, we pick the linearly decreasing schedule $\frac{1}{10^{5} + t}$ for SGD.
The reason for the specific constant $10^{5}$ in the denominator is that $10^{-5}$ is roughly the largest step size such that SGD does not diverge in its initial stage.

\subsection{Bayesian Logistic Regression}
On Mushroom, the step size of SGD is tuned from $\cbb{10^{-3}, 10^{-4}, 10^{-5}, 10^{-6}}$, while the step size of NGD is tuned from $\cbb{5 \cdot 10^{-1}, 10^{-1}, 10^{-2}, 10^{-3}}$.
On MNIST, the step size of SGD is tuned from $\cbb{10^{-5}, 10^{-6}, 10^{-7}}$, while the step size of NGD is tuned from $\cbb{10^{-1}, 10^{-2}, 10^{-3}}$.
Divergent curves (due to large step sizes) are not plotted in the graph.
We use $10$ samples from the variational distribution to estimate the stochastic gradient in every iteration.

Legends without the label ``(p)'' use the reparameterization trick to compute the stochastic gradient.
For SGD with the label ``(p)'', we use the Price theorem as follows.
First, observe the following relation between $\nabla_\Cv$ and $\nabla_{\Sigmav}$:
\begin{align*}
    \nabla_{\Cv} \Eb_{q(\zv)} \log p(\yv \mid \zv) & = 2 \cdot \nabla_{\Sigmav} \Eb_{q(\zv)} \sbb{\log p(\yv \mid \zv)} \cdot \Cv \\
    & = \Eb_{q(\zv)} \sbb{\nabla_{\zv}^2 \log p(\yv \mid \zv)} \cdot \Cv.
\end{align*}
To obtain a stochastic gradient estimate $\widehat\nabla_{\Cv} \Eb_{q(\zv)} \log p(\yv \mid \zv)$, replace the expectation with sample approximation.

\end{document}